\documentclass[twoside,11pt]{article}

\usepackage{amsmath,amsfonts}
\usepackage{epsf,epsfig,caption,subcaption,graphicx} 
\usepackage[ruled,boxed]{algorithm2e}
\usepackage{tikz}
\usepackage{natbib}
\usepackage{lastpage}
\usepackage{jmlr2e}

\usepackage{kotex}

\newcommand{\ech}{\color{black}  }    

\newtheorem{assumption}{Assumption} 

\def\var{\mbox{Var}}

\def\E{\mathbb{E}} 

\def\exp{\mbox{exp}}
\def\pa{\mbox{Pa}}
\def\ch{\mbox{Ch}}

\def\an{\mbox{An}}
\def\de{\mbox{De}}

\usepackage{mathtools}
\usepackage{etoolbox}

\DeclareFontFamily{U}{matha}{\hyphenchar\font45}
\DeclareFontShape{U}{matha}{m}{n}{
	<-6> matha5 <6-7> matha6 <7-8> matha7
	<8-9> matha8 <9-10> matha9
	<10-12> matha10 <12-> matha12
}{}
\DeclareSymbolFont{matha}{U}{matha}{m}{n}

\DeclareFontFamily{U}{mathx}{\hyphenchar\font45}
\DeclareFontShape{U}{mathx}{m}{n}{
	<-6> mathx5 <6-7> mathx6 <7-8> mathx7
	<8-9> mathx8 <9-10> mathx9
	<10-12> mathx10 <12-> mathx12
}{}
\DeclareSymbolFont{mathx}{U}{mathx}{m}{n}

\DeclareMathDelimiter{\vvvert} {0}{matha}{"7E}{mathx}{"17}%

\DeclarePairedDelimiterX{\normiii}[1]
{\vvvert}
{\vvvert}
{\ifblank{#1}{\:\cdot\:}{#1}}

\firstpageno{1}

\begin{document}
	
	\title{ 
		Bayesian Approach to Linear Bayesian Networks 
	}
	
	\author{\name Seyong Hwang \email seyong981121@gmail.com \\
		\addr Department of Statistics\\
		Seoul National University \\
		Seoul, 08826, South Korea
		\AND
		\name Kyoungjae Lee \email leekjstat@gmail.com \\
		\addr Department of Statistics\\
		Sungkyunkwan University \\
		Seoul, 03063, South Korea
		\AND		
		\name Sunmin Oh \email sunmin01111@gmail.com \\
		\addr Department of Statistics\\
		Seoul National University\\
		Seoul, 08826, South Korea
		\AND
		\name Gunwoong Park \email gw.park23@gmail.com \\
		\addr Department of Statistics\\
		Seoul National University \\
		Seoul, 08826, South Korea
	}
	\editor{  }
	
	\maketitle
	
	\begin{abstract}
		This study proposes the first Bayesian approach for learning high-dimensional linear Bayesian networks. The proposed approach iteratively estimates each element of the topological ordering from backward and its parent using the inverse of a partial covariance matrix. The proposed method successfully recovers the underlying structure when Bayesian regularization for the inverse covariance matrix with unequal shrinkage is applied. Specifically, it shows that the number of samples $n = \Omega( d_M^2 \log p)$ and $n = \Omega(d_M^2 p^{2/m})$ are sufficient for the proposed algorithm to learn linear Bayesian networks with sub-Gaussian and 4m-th bounded-moment error distributions, respectively, where $p$ is the number of nodes and $d_M$ is the maximum degree of the moralized graph. The theoretical findings are supported by extensive simulation studies including real data analysis. Furthermore the proposed method is demonstrated to outperform state-of-the-art frequentist approaches, such as the BHLSM, LISTEN, and TD algorithms in synthetic data.
	\end{abstract}
	
	
	\section{Introduction}

Bayesian networks are probabilistic graphical models that use a graph to represent variables of interest and their conditional dependence and causal relationships. Hence, Bayesian networks, especially Gaussian linear Bayesian networks, have been applied in various fields due to their popularity and usefulness. (e.g., \citealp{lauritzen1996graphical,spirtes2000causation,koller2009probabilistic,chickering2013bayesian}). 

Recent studies have investigated structure learning algorithms for (Gaussian) linear Bayesian networks (BNs), also referred to as linear structural equation models (SEMs). These algorithms can be categorized into three groups: (i) likelihood-based learning, (ii) inverse covariance matrix-based, and (iii) a node-wise regression-based algorithms. For example, \citet{peters2014identifiability} develops the likelihood-based greedy DAG search algorithm in large sample settings. \citet{loh2014high,ghoshal2018learning} provide the regularized inverse covariance matrix-based algorithms for high-dimensional sparse Gaussian linear BNs. \citet{chen2019causal,park2021learning2} develop the regularized-regression-based algorithms. Additionally, \citet{park2020identifiability,park2020identifiability2} propose the inverse covariance matrix and independence test-based uncertainty scoring (US) algorithms for low-dimensional settings, and \citet{park2021learning} develops the high-dimensional Gaussian linear BN learning algorithm by combining the graphical Lasso and the US algorithm. However, the study of learning algorithms under the Bayesian framework is less developed, due to the significant computational cost and the nascent status of this field. 

The main objective of this study is to develop a Bayesian approach for learning high-dimensional linear BNs using the principles of inverse covariance matrix-based approaches. Specifically, the proposed algorithm applies an iterative step of inferring element-wise ordering and its parents, where both problems are effectively addressed using Bayesian regularization for inverse covariance matrix estimation with unequal shrinkage (BAGUS). In addition, the theoretical guarantees on the proposed algorithm are shown, that is the numbers of samples $n = \Omega(d_M^2 \log p)$ and $n = \Omega(d_M^2p^{2/m})$ are sufficient for the proposed algorithm to recover the underlying structure, for sub-Gaussian and 4m-th bounded-moment linear BNs, respectively, where $p$ is the number of nodes and $d_M$ is the maximum degree of the moralized graph.

The theoretical findings of this study are heuristically confirmed through extensive simulation studies. The proposed algorithm consistently recovers the underlying graph with sample complexities of $n = \Omega(d_M^2 \log p)$ and $n = \Omega(d_M^2p^{2/m})$ for sub-Gaussian and 4m-th bounded-moment linear BNs, respectively. Furthermore, the proposed Bayesian approach is compared to state-of-the-art frequentist approaches, such as BHLSM~\citep{park2021learning2}, LISTEN~\citep{ghoshal2018learning}, TD~\citep{chen2019causal}, and US~\citep{park2020identifiability2} algorithms. Finally, we demonstrate through online shopping mall order amount data that the proposed algorithm is well-suited for estimating the relationship between the sales of each product.

The remainder of this paper is structured as follows. Section~\ref{SecGSEM} summarizes some necessary notations, explains basic concepts of a linear BN, and discusses the identifiability conditions and existing learning algorithms for linear BNs. Section~\ref{SecAlgo} introduces the proposed algorithm for high-dimensional linear BNs. Section~\ref{SecTheo} provides theoretical guarantees of the proposed algorithm and illustrates specific examples. Section~\ref{SecComparison} compares the proposed algorithm with other frequentist high-dimensional linear BN learning algorithms. Section~\ref{SecNume} evaluates the proposed algorithm and compares it with state-of-the-art algorithms in various simulation settings. Section~\ref{SecReal} demonstrates that the proposed algorithm performs well for estimating relationships between sales of each product of an online mall. Lastly, Section~\ref{SecFuture} offers a discussion and suggests a future work.

\section{Background}\label{SecGSEM}

In this section, we will first introduce some necessary notations and definitions for linear Bayesian networks (BNs), also known as linear structural equation models (SEMs). Then, we will discuss previous relevant works.


\subsection{Bayesian Network}

A Bayesian network (BN) is a probabilistic graphical model that represents a joint probability distribution over a set of random variables using a directed acyclic graph (DAG) and a set of conditional probability distributions. DAG $G = (V, E)$ consists of a set of nodes $V = \{1, 2, ... , p\}$ and a set of directed edges $E \subset V \times V$ with no directed cycles. A directed edge from node $j$ to $k$ is denoted by $(j,k)$ or $j \rightarrow k$. The set of \emph{parents} of node $k$, denoted by $\pa(k)$, consists of all nodes $j$ such that $(j,k) \in E$. In addition, the set of \emph{children}, denoted by $\ch(j)$, consists of all nodes $k$ such that $(j,k) \in E$. If there is a directed path from node $j$ to node $k$, then $k$ is called a \emph{descendant} of $j$, and $j$ is called an \emph{ancestor} of $k$. The sets of all descendants and ancestors of node $k$ are denoted by $\de(k)$ and $\an(k)$, respectively. An important property of DAGs is that there exists (topological) \emph{ordering} $\pi = (\pi_1,\pi_2, ...., \pi_p)$ of a directed graph that represents the directions of edges such that for every directed edge $(j, k) \in E$, $j$ comes before $k$ in the ordering. Hence, recovering a graph can be decomposed into learning the ordering and the presence of edges.

In Bayesian network, we consider a random vector $X := (X_j)_{j \in V}$ with a probability distribution taking values in sample space $\mathcal{X}_{V}$ over the nodes in $G$. For any subset $S$ of $V$, let $X_{S} :=(X_j)_{j \in S}$ and $\mathcal{X}_{S} := \times_{j \in S} \mathcal{X}_{j}$ where $\mathcal{X}_{j}$ is the sample space of $X_j$. For any node $j \in V$, $\Pr(X_j \mid X_{S})$ denotes the conditional distribution of the variable $X_j$ given a random vector $X_{S}$. Then, a DAG model has the following joint probability density function:
\begin{equation*}
	\label{eq:factorization}
	\Pr(X_1, X_2,..., X_p) = \prod_{j=1}^{p} \Pr(X_j \mid X_{\pa(j)}),
\end{equation*}
where $\Pr(X_j \mid X_{\pa(j)})$ is the conditional distribution of $X_j$ given its parent variables $X_{\pa(j)} =(X_k)_ {k \in \pa(j)}$.

This study considers $n$ independent and identically distributed (i.i.d.) samples $X^{1:n} := ( X^{(i)} )_{i=1}^{n}$ from a given Bayesian network where $X^{(i)} := ( X_j^{(i)} )_{j = 1}^{p}$ is a $p$-variate random vector. The notation $\widehat{\cdot}$ denotes an estimate based on samples $X^{1:n}$. This study assumes that the moralized graph is sparse, meaning that the Markov blanket of each node is small. Additionally, it assumes causal sufficiency, meaning that $X_{\pa(j)}$ is the only source of confounding for $X_j$.

\subsection{Linear Bayesian networks}

\label{SecLSEM}

A linear Bayesian network, also known as a linear structural equation model, is a type of Bayesian network (BN) where the joint distribution is defined by linear structural equations. The model can be expressed in a matrix equation as follows:
\begin{equation}
	\label{eq:LinearSEM}
	(X_1, X_2,..., X_p)^\top =  B (X_1, X_2,..., X_p)^\top + (\epsilon_1,\epsilon_2,  ... , \epsilon_p)^\top,
\end{equation}
where $B \in \mathbb{R}^{p \times p}$ is an edge weight matrix with each element $[B]_{j,k} = \beta_{k,j}$, representing the linear weight of an edge from $X_k$ to $X_j$. In addition, $(\epsilon_j)_{j \in V}$ are independent error random variables with $\E(\epsilon_j) = 0$ and $\var(\epsilon_j) = \sigma_j^2>0$. 

Then, the covariance matrix of $X$ and its inverse, $\Sigma$ and $\Omega$, respectively, can be calculated as:
\begin{equation}
	\label{eq:MatrixSEMSigma}
	\Sigma  = (I_p - B)^{-1} \Sigma^{\epsilon} (I_p - B)^{-\top}, \text{ and}  \quad \Omega = (I_p - B)^{\top} (\Sigma^{\epsilon})^{-1} (I_p - B),
\end{equation}
where $I_{p} \in \mathbb{R}^{p \times p}$ is the identity matrix, and $\Sigma^{\epsilon} = \mbox{diag}(\sigma_{1}^2,\sigma_{2}^2, ..., \sigma_{p}^2)$ is the covariance matrix for the independent errors.

Since the inverse covariance matrix is a function of edge weights, many existing algorithms apply the inverse covariance matrix to recover the structure, i.e., the support of $B$. More precisely, the algorithms learn the models under the following \textit{uncertainty level} conditions. 

\begin{lemma}[Identifiability Conditions in Theorem 4 of \citealp{park2020identifiability2}]
	\label{lem:identifiability}
	Let $X$ 
	be generated from a linear BN~\eqref{eq:LinearSEM} with DAG $G$ and true ordering $\pi$. In addition, suppose that $\Sigma$ is the true covariance matrix. Then, DAG $G$ is uniquely identifiable if either of the two following conditions is satisfied: Consider any node $j = \pi_r \in V$, $k \in \de(j)$, $\ell \in \an(j)$, $S_r = \{\pi_1, ..., \pi_{r-1} \}$, and $T_r = V \setminus \{ \pi_{r+1}, ..., \pi_{p} \} $ in which $\pi_0 = \pi_{p+1} = \emptyset$. 
	\begin{itemize}
		\item[(1)] Forward selection~\citep{park2020identifiability}:
		$$
		\frac{1}{
			[(\Sigma_{S_r \cup \{j\}, S_r\cup \{j\}})^{-1}]_{j,j} } = \sigma_j^2 < \sigma_k^2 + 
		\sum_{k' \in \an(k) \setminus \{\pi_1, ..., \pi_{r-1} \} } \beta_{k' \to k}^2 \sigma_{k'}^2 
		= \frac{1}{
			[(\Sigma_{S_r \cup \{k\}, S_r\cup \{k\}})^{-1}]_{k,k} } 
		$$
		where $\beta_{k' \to k}$ is the sum over products of coefficients along each directed path from $k'$ to $k$. In addition, $\Sigma_{A,A}$ is the $|A| \times |A|$ sub-matrix of $\Sigma$ corresponding to variables $X_{A}$. Finally, $[ \Sigma ]_{k,k}$ is the diagonal entry corresponding to variable $X_k$. 
		
		\item[(2)] Backward selection \citep{ghoshal2018learning}: 
		$$
		[(\Sigma_{T_r, T_r})^{-1}]_{j,j} = \frac{1}{\sigma_j^2} <   \frac{1}{\sigma_{\ell}^2} + \sum_{\ell' \in \ch(\ell) \setminus \{\pi_{r}, ..., \pi_{p}\} } \frac{ \beta_{\ell, \ell'}^2 }{ \sigma_{\ell'}^2} = [(\Sigma_{T_r, T_r})^{-1}]_{\ell,\ell}.
		$$
	\end{itemize}
\end{lemma}

It is straightforward that if the error variances are the same, both identifiability conditions hold. Hence, in many areas, these identifiable models are acceptable and widely used. For example, the assumption of exact same error variances is used for applications with variables from a similar domain, such as spatial or time-series data.

It is also important to note that these identifiability conditions do not rely on any specific distribution, such as Gaussian. However, in order to achieve high-dimensional consistency, two types of linear BNs are considered, which are concerned with the tail conditions of error distributions. The first type is sub-Gaussian linear BN, which is a natural extension of Gaussian linear BNs, in which each error variable is sub-Gaussian. That is, $\epsilon_j / \sqrt{\var (\epsilon_j) }$ is sub-Gaussian with parameter $s_{max}^2$. The second type is bounded-moment linear BN, which is defined as the linear BN with errors having a bounded moment. Specifically, $\max_{j \in V} \E{ (\epsilon_j / \sqrt{\var (\epsilon_j) } )^{4m} } \leq K_{m}$, where $K_m >0$ only depends on $m$.

In the linear BN, each variable $X_j$ can be expressed as a linear combination of independent errors corresponding to its ancestors:
$$
X_j = \sum_{ k \in \pa(j) }\beta_{k,j} X_{k} + \epsilon_j = \sum_{k \in \an{(j)}} \beta_{k \rightarrow j} \epsilon_k + \epsilon_j,
$$
where $\beta_{k \to j}$ is the sum over products of coefficients along directed paths from $k$ to $j$. Hence, if the error variables have a sub-Gaussian or a bounded-moment property, then $X_j$ also satisfies a sub-Gaussian or a bounded-moment property. These two types of linear BNs can be expressed as follows.
\begin{itemize}
	\item Sub-Gaussian linear BN: 
	For any $j \in V$, 
	$X_j$ is sub-Gaussian with proxy parameter $s_{\max}^2 [\Sigma]_{j,j}$ for some constant $s_{\max}>0$, which means that 
	\begin{equation}\label{subGaussian}
		\E\{ \exp (t X_j) \}   \le  \exp (s_{\max}^2 [\Sigma]_{j,j} t^2/ 2 )  \text{ for all } t\in\mathbb{R} .
	\end{equation}
	
	\item $4m$-th bounded-moment linear BN: For any $j \in V$, 
	$X_j$ has $4m$-th bounded-moment for some positive integer $m$ and 
	some constant $K_{\max}>0 $  
	such that 
	\begin{equation}\label{4mbdd}
		\E\{ (X_j)^{4m} \} \leq K_{\max} [\Sigma]_{j,j}^{2m} .
	\end{equation}
\end{itemize}

\subsection{Frequentist Approaches to Linear BNs}

\label{SecExisting}

This section provides a brief review of recent frequentist methods for learning linear BNs. One popular approach is based on regression. For example, \citet{park2020identifiability} applies a standard regression and a conditional independence test to learn a low-dimensional Gaussian linear BN. It estimates the ordering using the variance of residuals, and subsequently infers the directed edges using conditional independence tests. \citet{chen2019causal} develop the $\ell_0$- and $\ell_1$-regression-combined TD algorithm for a sub-Gaussian linear BN. The algorithm estimates the ordering using the best-subset-regression, and then infers the edges using a $\ell_1$-regularized approach. The TD algorithm requires a sample size of $n = \Omega(q^2 \log p)$ for accurate ordering estimation, where $q$ is the predetermined upper bound of the maximum indegree.

Similarly, \citet{park2021learning2} develops the $\ell_1$-regression-based linear BN learning (LSEM) algorithm with sample complexities of $n = \Omega(d_M^4 \log p)$ and $n = \Omega(d_M^4 p^{2/m} )$ for sub-Gaussian and $4m$-th bounded-moment linear BNs, respectively. Finally, \citet{gao2022optimal} proposes the best-subset-selection-based optimal learning algorithm for Gaussian linear BNs. Its sample complexity is optimal $n = \Omega(d_{in} \log \frac{p}{d_{in}})$ under the known maximum indegree $d_{in}$.

Another popular approach is based on the inverse covariance matrix. These algorithms estimate the last element of the ordering using the diagonal entries of the inverse covariance matrix, which can be estimated by any inverse covariance matrix estimators, such as graphical Lasso and CLIME. These algorithms then determine its parents with non-zero entries on its row of the inverse covariance matrix. After eliminating the last element of the ordering, this procedure is repeated until the graph is fully estimated. \citet{loh2014high} and \citet{ghoshal2018learning} apply graphical Lasso and CLIME, respectively, for inverse covariance matrix estimation, and prove that their algorithms require sample sizes of $n = \Omega(d_M^4 \log p)$ and $n = \Omega(d_M^4 p^{2/m} )$, when the error distributions are sub-Gaussian and $4m$-th bounded-moment, respectively.

In summary, the existing algorithms for learning linear BNs mainly use frequentist methods such as OLS, Lasso, graphical Lasso, and CLIME. However, a Bayesian framework structure learning algorithm has not yet been explored due to its heavy computational cost, model complexity, and choice of prior. Therefore, this study proposes an inverse covariance matrix-based Bayesian algorithm for high-dimensional linear BNs, using a provable and scalable Bayesian approach for inverse covariance matrix estimation.

\subsection{Bayesian Approaches to Undirected Graphical Models}

\label{SecBAGUS}

Several Bayesian methods have been proposed for estimating high-dimensional sparse Gaussian undirected graphical models, that is equivalent to high-dimensional sparse inverse covariance matrices. For example, \citet{roverato2000cholesky} suggests using the $G$-Wishart distribution as a prior. \citet{lee2021bayesian} proves theoretical properties such as posterior convergence rate and graph selection consistency when a carefully chosen prior is used for the graph $G$. However, since its normalizing constant has a closed form only when the graph $G$ is decomposable, posterior inference for a general graph $G$ requires a computationally expensive Markov chain Monte Carlo (MCMC) algorithm.	

As alternatives, spike-and-slab and its variants as priors for sparse inverse covariance matrix have been suggested \citep{wang2015scaling, banerjee2015bayesian, gan2019bayesian}. 
Due to the positive definite constraint, the resulting prior usually contains an unknown normalizing constant. Hence, block Gibbs samplers have been suggested for efficient posterior inference.


In this study, we concentrate on the Bayesian regularization for graphical models with unequal shrinkage (BAGUS) approach proposed by \citet{gan2019bayesian}. The off-diagonal entries of the inverse covariance matrix $\Omega \in \mathbb{R}^{p\times p}$ are modeled using a spike-and-slab prior as follows (see details in \citealp{rovckova2018spike}): for any $1\le j < k \le p$,
\begin{align}\label{BAGUS_prior}
	\pi([\Omega]_{j,k}) = \frac{\eta}{2 \nu_1} \exp \left( -\frac{|[\Omega]_{j,k}|}{ \nu_1 } \right) + \frac{1 - \eta }{ 2\nu_0} \exp \left( -\frac{|[\Omega]_{j,k}|}{ \nu_0 } \right), \text{ s.t. }
	[\Omega]_{k,j} = [\Omega]_{j,k}
\end{align}
where $\nu_1>\nu_0 >0$ are  scaling hyper-parameters, and $0<\eta<1$ is a hyper-parameter that reflects the prior belief about the proportion of signals. Note that the spike-and-slab prior in Equation~\eqref{BAGUS_prior} consists of two Laplace distributions. The first component (slab), with relatively large scale $\nu_1$, captures large signals, whereas the second component (spike), with small scale $\nu_0$, shrinks small noises to zero. In addition, the BAGUS method assumes the following exponential prior for the diagonal entries of $\Omega$, which is defined as follows: for any $1\le j \le p$, 
$$
\pi( [\Omega]_{j,j}) = \tau \exp \left( - \tau [\Omega]_{j,j} \right), \text{ s.t. } [\Omega]_{j,j} > 0,
$$
where $\tau>0$ is a hyper-parameter.

To define a prior distribution for inverse covariance matrices, the BAGUS method applies the following prior that is restricted to a subset of positive definite matrices:
$$
\pi( \Omega ) = \prod_{j < k} \pi([\Omega]_{j,k}) \prod_{j} \pi( [\Omega]_{j,j}) 1(\Omega \succ 0 , \, \|\Omega\|_2 \leq B_0),
$$
where $B_0>0$ is a constant and $1(\cdot)$ is an indicator function. Additionally, $\Omega \succ 0$ means that $\Omega$ is positive definite, and $\|\Omega\|_2 = \sup_{x \in \mathbb{R}^p ,\|x\|_2=1 } \|\Omega x \|_2$ is the spectral norm of $\Omega$.

Then, the BAGUS method returns the maximum a posteriori (MAP) estimator, say $\widehat{\Omega}$, which maximizes the  posterior distribution using the expectation-maximization (EM) algorithm:
\begin{align}
	\label{eq:BAGUS}
	\widehat{\Omega} := \underset{ \substack{\Omega \succ 0 , \\ \| \Omega \|_2 \leq B_0} } {\mbox{arg max} }  \frac{n}{2} \left\{ \mbox{tr}( \widehat{\Sigma} \Omega) - \log \det (\Omega) \right\} - \sum_{j <k} \log \left( \frac{\eta}{2 \nu_1} e^{ -\frac{|[\Omega]_{j,k}|}{ \nu_1 } } + \frac{1 - \eta }{ 2\nu_0} e^{-\frac{|[\Omega]_{j,k}|}{ \nu_0 } } \right) + \tau \sum_{j=1}^p [\Omega]_{j,j},
\end{align}
where $\widehat{\Sigma}$ is the sample covariance matrix. 

\citet{gan2019bayesian} shows that the EM algorithm has a computational complexity of $O(p^3)$. Hence, the BAGUS method is much more tractable compared to other Bayesian methods based on the MCMC algorithm. Moreover, \citet{gan2019bayesian} establishes 
convergence rates of the MAP estimator 
in high-dimensional settings under regularity conditions on hyperparameters and the true inverse covariance matrix. Specifically, they shows that the BAGUS method consistently estimates the true inverse covariance matrix if the sample size scales as $n = \Omega(d^2 \log p)$, where $d$ is the maximum cardinality of nonzero elements in each column of the true inverse covariance matrix, which is equivalent to the maximum degree of the moralized graph in the language of DAG models.

\citet{gan2019bayesian} further proposes a consistent support recovery procedure based on the posterior inclusion probability. The spike-and-slab prior in Equation~\eqref{BAGUS_prior} can be written as the following hierarchical prior:
\begin{align*}
	[\Omega]_{j,k} \mid  r_{j,k} =0 &\sim \mbox{Laplace}(\nu_0), ~~
	[\Omega]_{j,k} \mid  r_{j,k} =1 \sim \mbox{Laplace}(\nu_1), ~~
	r_{j,k} \sim \mbox{Bern}(\eta)  ,
\end{align*}
where $\mbox{Laplace}(\nu)$ is the probability density function of the Laplace distribution with the scale parameter $\nu$.
The binary variable $r_{j,k}$ indicates whether $[\Omega]_{j,k}$ is zero or not; hence, we can conduct variable selection based on the posterior inclusion probability:
\begin{align*}
	\Pr( r_{j,k} =1 \mid  X^{1:n} )  &=  \int \Pr( r_{j,k} =1 \mid [\Omega]_{j,k} ) \pi([\Omega]_{j,k} \mid X^{1:n} ) d [\Omega]_{j,k} .
\end{align*}
The above posterior inclusion probability can be estimated based on the MAP estimator $\widehat{\Omega}$ as follows:
\begin{align*}
	p_{j,k} &:= \Pr(r_{j,k} =1 \mid  [\widehat{\Omega}]_{j,k} )  \\
	&= \frac{ \eta / (2\nu_1) \, \exp \big(- | [\widehat{\Omega}]_{j,k}| /\nu_1 \big)  }{\eta / (2\nu_1) \, \exp \big(- | [\widehat{\Omega}]_{j,k}| /\nu_1 \big) + (1-\eta) / (2\nu_0) \, \exp \big(- | [\widehat{\Omega}]_{j,k}| /\nu_0 \big) } .
\end{align*}
\cite{gan2019bayesian} shows that, under regularity conditions, the estimated support $\hat{S} = \{(j,k) : p_{j,k} \ge T \}$ is consistent for any threshold $0<T<1$.

\section{Algorithm}\label{SecAlgo}

This section presents a Bayesian approach for learning high-dimensional linear BNs. The proposed algorithm combines the BAGUS method and the principle of inverse covariance matrix-based linear BN learning approaches. Specifically, at the first step, it estimates the inverse covariance matrix for all variables. Then, the last element of the ordering is chosen with the smallest diagonal entry using the backward selection condition in Lemma~\ref{lem:identifiability}. Subsequently, its parents are determined by the indices with inclusion probabilities greater than a given threshold $0<T<1$, which represents whether an edge weight is zero or not in Equation~\eqref{eq:LinearSEM}. 
In the next step, the inverse covariance matrix is estimated using only the remaining variables except the last element of the ordering.
The proposed algorithm estimates the next element of the ordering and its parents with the same method. It iterates this procedure until the complete graph structure is inferred. The detailed process of the proposed algorithm is summarized in Algorithm \ref{OurAlgorithm}.

\setlength{\algomargin}{0.5em}
\begin{algorithm}[ht]
	\caption{\bf Bayesian Linear BN Learning Algorithm \label{OurAlgorithm}}
	\SetKwInOut{Input}{Input}
	\SetKwInOut{Output}{Output}
	\SetKwInOut{Return}{Return}
	\Input{ $n$ i.i.d. samples $X^{1:n}$ and a threshold $0<T<1$. }
	\Output{ Estimated graph structure, $\widehat{G} = (V, \widehat{E})$. }
	\BlankLine
	Set $\widehat{\pi}_{p+1} = \emptyset$ \; 
	\For{$r = \{1,2,...,p-1\}$}{
		Set $S(r) = V \setminus \{ \widehat{\pi}_{p+2-r}, ...,  \widehat{\pi}_{p+1} \} $\;
		Estimate inverse covariance matrix $\widehat{\Omega}^{(r)}$ for $X_{S(r)}$ using the Bayesian approach in Equation~\eqref{eqn:InvCov}\;
		Estimate $\widehat{\pi}_{p+1-r} = \arg \min_{j \in S(r) } [\widehat{\Omega}^{(r)}]_{j,j}\ech$\;
		Estimate $\widehat{\pa}(\widehat{\pi}_{p+1-r}) = \{k \in S(r) \setminus \{\widehat{\pi}_{p+1-r}  \}  : 
		p_{\widehat{\pi}_{p+1-r},k}^{(r)}  \ge T  
		\}$, where  $p^{(r)}_{\widehat{\pi}_{p+1-r},k} := \Pr(r_{\widehat{\pi}_{p+1-r},k}=1 \mid [\widehat{\Omega}^{(r)}]_{\widehat{\pi}_{p+1-r},k} )$
	}
	\Return{Estimated edge set, $\widehat{E} =  \cup_{r \in \{1,2, ... ,p-1\} } \{ (k, \widehat{\pi}_{p+1-r}) : k \in \widehat{\pa}(\widehat{\pi}_{p+1-r})\} $.}
\end{algorithm}

In Algorithm~\ref{OurAlgorithm}, the $r$-th iteration of the algorithm is first conducted by calculating the following MAP estimator of the inversion of the partial covariance matrix based on the BAGUS method:
\begin{align}
	\label{eqn:InvCov}
	\widehat{\Omega}^{(r)} := &\underset{ \substack{ \Theta \succ 0 , \|  \Theta \|_2 \leq B_0 , \nonumber \\ 
			\Theta \in \mathbb{R}^{(p+1-r) \times (p+1-r)}} }{\mbox{arg max} }  \frac{n}{2} \left\{   \mbox{tr}(\widehat{\Sigma}^{(r)}  \Theta ) - \log \det ( \Theta) \right\} -\\ 
	&\sum_{j <k,j,k\in S(r)} \log \left( \frac{\eta}{2 \nu_1} e^{ -\frac{| [\Theta]_{j,k} |}{ \nu_1 } } + \frac{1 - \eta }{ 2\nu_0} e^{-\frac{| [\Theta]_{j,k}|}{ \nu_0 } } \right) + \tau \sum_{j \in S(r)}  [\Theta]_{j,j},
\end{align}
where $\widehat{\Sigma}^{(r)}$ is a sample covariance matrix for $X_{S(r)}$ in which $S(r) = V \setminus \{  \widehat{\pi}_{p+2-r} ,\ldots, \widehat{\pi}_{p+1} \}$ and $\widehat{\pi}_{p+1} =\emptyset$.  Then, it determines $\widehat{\pi}_{p+1-r}$ with the smallest diagonal entries of the estimated inverse covariance matrix $\widehat{\Omega}^{(r)}$. 
Finally, it determines the parents of node $j = \widehat{\pi}_{p+1-r}$ such that the posterior probability of non-zero $[\widehat{\Omega}^{(r)}]_{j,k}$ is greater or equal to the pre-specified threshold.
We discuss the required conditions on the threshold in Section~\ref{SecTheo}. 

When recovering the ordering, it is recommended to choose a small value of the tuning parameter $\tau$ in Equation~\eqref{eqn:InvCov}, which is the inverse scale for diagonal entries. This is because the diagonal entries of the inverse covariance matrix do not need to be shrunk to zero. Additionally, small $\nu_0$ for the spike part and large scale $\nu_1$ for the slab part are recommended to achieve a sparse estimated graph. The rationale behind this is that the most off-diagonal entries of the inverse covariance matrix should be shrunk to zero while keeping the large signals, when recovering the parents in a sparse graph setting.

It is emphasized that the proposed method is based on the estimation of the edges of the moralized graph instead of the DAG.
Specifically, the proposed method estimates the uncertainty score for the ordering estimation using the posterior distribution of the moralized graph. 
Subsequently, it recovers the parents using the posterior distribution of the moralized graph again for better computational efficiency. 
Additionally, it would make more sense to exploit the posterior distribution of the directed graph when recovering the directed edges given the ordering, as shown in previous studies \citep{viinikka2020towards}. 

Regarding computational complexity, the bottleneck of the $r$-th iteration of the proposed method is the estimation of an inverse covariance matrix with the size of $(p+1-r) \times (p+1-r)$ using the BAGUS method. According to \citet{gan2019bayesian}, its computational cost is $O((p+1-r)^3)$. Since there are $p-1$ iterations, the computational complexity of the proposed method is $\sum_{j=2}^{p} O(j^3) = O(p^4)$. Hence, the proposed method has polynomial computational complexity in the number of nodes.

\section{Theoretical Results}

\label{SecTheo}

This section provides the statistical results for Algorithm~\ref{OurAlgorithm} in learning high-dimensional linear BNs with sub-Gaussian and $4m$-th bounded-moment errors. Specifically, we present the required assumptions and the theoretical results for the consistent estimation of both the ordering and directed edges. The main results are expressed in terms of the triple $(n,p,d_M)$, where $n$ is the number of samples, $p$ is the number of nodes, and $d_M$ is the maximum degree of the moralized graph. For ease of notation, let $X_{\pi_{1:r}} = (X_{\pi_1}, X_{\pi_2},...,X_{\pi_r})$.

We assume some prevalent constraints on the linear BN~\eqref{eq:LinearSEM} with both sub-Gaussian and 4m-th bounded-moment error distributions. (e.g.,~\citealp{ghoshal2018learning,chen2019causal,park2021learning2,gao2022optimal})

\begin{assumption}[Dependency Assumption]
	\label{assu:eigen}
	There exists a positive constant $k_1$ and $k_2 > 0$ such that
	\begin{align*}
		\Lambda_{\min}\left( \Sigma \right) \geq k_1 \quad \text{and} \quad \max_{j \in \{1,2,...,p\}} [\Sigma]_{j,j} < k_2
	\end{align*}
	where $\Lambda_{\min}(A)$ denotes the minimum eigenvalue of matrix $A$.
\end{assumption}

\begin{assumption}[Minimum Gap Assumption]
	\label{assu:condvar} 
	For any $r \in \{2,3,...,p\}$ and $j = \pi_r$ and $\ell \in \an(j)$, there exists a positive constant $\tau_{\min} > 0$ such that 
	$$
	-\frac{1}{\sigma_{j}^2} + \frac{1}{\sigma_{\ell}^2} + \sum_{\ell' \in \ch(\ell) \setminus \{\pi_{r}, ..., \pi_{p}\} } \frac{ \beta_{\ell, \ell'}^2 }{ \sigma_{\ell'}^2} > \tau_{\min}.
	$$
\end{assumption}

\begin{assumption}[Minimum Signal Assumption]
	\label{assu:min}
	For any $r \in \{1,2,...,p-1\}$, there exists a positive constant $\theta_{\min} > 0$ such that 
	\begin{align*}
		\min_{(k, j) \in E^{(r)} } 	|\Omega^{(r)}_{j,k}|  
		> \theta_{\min},
	\end{align*}
	where $E^{(r)}$ and $\Omega^{(r)}$ is an edge set and an inverse covariance matrix of $\{\pi_1,\pi_2,...,\pi_{p+1-r}\}$, respectively.
\end{assumption}

Assumption~\ref{assu:eigen} is necessary because the proposed algorithm relies on the inverse covariance matrix, which requires an invertible covariance matrix. Assumption~\ref{assu:condvar} is a sample version of the backward selection condition in Lemma~\ref{lem:identifiability}, which ensures that the difference between uncertainty scores is large enough to identify the correct ordering. Furthermore, Assumption~\ref{assu:min} ensures that, for any $r \in \{1,2,...,p-1\}$, the $\pi_{p+1-r}$-th row of non-zero entries of the inverse covariance matrix for $X_{\pi_{1:(p+1-r)}}$ are sufficiently far away from zero. Since the edge weight is proportional to $[\Omega^{(r)}]_{\pi_{p+1-r}, k}$ from Equation~\eqref{eq:MatrixSEMSigma}, the assumption guarantees that each non-zero edge weight is sufficiently large.

For learning high-dimensional models, we require the following related conditions on the covariance and inverse covariance matrix. For ease of notation, let $S = \{(i,j):[\Omega]_{i,j} \neq 0\}$ and $S^{(r)} = \{(i,j):[\Omega^{(r)}]_{i,j} \neq 0\}$. We also define $M_{\Sigma} = \normiii{\Sigma}_\infty$, $M_{\Gamma} = \normiii{(\Omega \otimes \Omega)_{SS}}_\infty$ and $M_{\Gamma^{(r)}} = \normiii{(\Omega^{(r)} \otimes \Omega^{(r)})_{S^{(r)}S^{(r)}}}_\infty$ for any $r \in \{1,2,...,p-1\}$, where $\normiii{\cdot}_\infty$ and $\otimes$ denote the $\ell_{\infty}/\ell_{\infty}$ operator norm and Kronecker product, respectively. 
Lastly, suppose that $M_{\Gamma_{\max}} = \max_{r \in \{1,2,...,p-1\}} M_{\Gamma^{(r)}}$ and $M_{\Gamma_{\min}} = \min_{r \in \{1,2,...,p-1\}} M_{\Gamma^{(r)}}$.

\begin{assumption}
	\label{assu:spar}
	There exists positive constants $M_1 > 0$ and $M_2 > 0$ such that 
	\begin{align*}
		M_{\Sigma} < M_1 \quad \text{ and } \quad M_{\Gamma_{max}} < M_2.
	\end{align*}
\end{assumption}

Assumption~\ref{assu:spar} constrains the $\ell_{\infty}/\ell_{\infty}$ operator norm of the inverse covariance matrix, whose support is related to the moralized graph, thereby controlling the sparsity of the moralized graph. Specifically, this assumption is particularly favorable for a DAG with a sparse moralized graph, such as a chain graph. In contrast, it is harsh for a DAG with a dense moralized graph, such as a star graph (for further details, see Section \ref{Secillustration}). The similar assumption is also employed in the graphical Lasso method proposed in \citet{ravikumar2011high} for learning the inverse covariance matrix. \citet{ravikumar2011high} constrains the $\ell_{\infty}/\ell_{\infty}$ operator norm of $\Omega$ to be constant whereas the proposed method requires restraint for all $\Omega^{(r)},\text{ }r\in\{1,2,...,p\}$ to estimate the inverse covariance matrix for each iteration.

Armed with Assumptions~\ref{assu:eigen},~\ref{assu:condvar},~\ref{assu:min} and~\ref{assu:spar}, the main result is achieved, which shows that the proposed algorithm can consistently learn a high-dimensional linear BN.

\begin{theorem}[Consistency]
	\label{thm001}
	Consider a linear BN~\eqref{eq:LinearSEM} with sub-Gaussian and $4m$-th bounded-moment errors. Suppose that Assumptions~\ref{assu:eigen}, \ref{assu:condvar}, \ref{assu:min}, and \ref{assu:spar} are satisfied. Additionally, suppose that there exists constants $C_1>0$, $C_2 > C_3>0$, $\epsilon_1 >0$ such that 
	\begin{align*}
		C_3\epsilon_1 \le \frac{k_1^2p}{2}, \text{ and }(C_1+C_3) < \frac{1}{4 M_{\Gamma_{max}} } \min\left\{ \tau_{\min} , 2 \theta_{\min} d, k_1^2,
		\frac{2}{3M_{\Sigma}}, 
		\frac{2}{3M_{\Gamma_{max}}M_{\Sigma}^3}
		\right\}.
	\end{align*} 	
	Finally, suppose that the prior hyper-parameters $(v_0, v_1, \eta, \tau)$, threshold parameter $T$, and the spectral norm bound $B_0$ satisfy
	\begin{align*}
		\frac{1}{v_1} &= \frac{C_3}{1+\epsilon_1}\frac{n}{p}, \quad
		\frac{1}{v_0} > C_4\frac{n}{p}, \quad
		\frac{v_1^2(1-\eta)}{v_0^2\eta} \le \epsilon_1\exp\left\{ 2(C_2-C_3)M_{\Gamma_{min}}(C_4-C_3) \frac{n}{p^2}\right\}, \\[3mm] 
		\tau & \le C_3\frac{n}{2p}, \quad \log\left(\frac{T}{1-T}\right) \in \log\left(\frac{\nu_0\eta}{\nu_1(1-\eta)}\right)+ \left(0, \left(\theta_{\min}-2(C_1+C_3)M_{\Gamma}\frac{1}{d}\right)\left(\frac{1}{\nu_0}-\frac{1}{\nu_1}\right)\right), \\[3mm] \text{ and } &\frac{1}{k_1} + 2(C_1+C_3)M_{\Gamma_{max}} < B_0 < \sqrt{2nv_0},
	\end{align*} 
	where $C_4 = C_1+M_{\Sigma}^22(C_1+C_3)M_{\Gamma_{max}}+6(C_1+C_3)^2M_{\Gamma_{max}}^2M_{\Sigma}^3$.
	Then, Algorithm~\ref{OurAlgorithm} correctly estimates the graph with high probability as follows:
	\begin{itemize}
		\item For a sub-Gaussian linear BN,
		\begin{align*}
			\Pr \left( \widehat{G} = G \right) \ge 1-4p^2  \exp\left( - \frac{ C_1^2  }{128(1+4 s_{\max}^2)\max_{j}\left( [\Sigma]_{j,j}\right)^2}\frac{n}{(d_M+1)^2}\right).
		\end{align*}
		\item For a $4m$-th bounded-moment linear BN,
		\begin{align*}
			\Pr \left( \widehat{G} = G \right) \ge 1-4p^2\frac{2^{2m}\max_{j}\left( [\Sigma]_{j,j}\right)^{2m}C_m(K_{\max}+1)}{C_1^{2m}} \frac{(d_M+1)^{2m}}{n^m}.
		\end{align*}
	\end{itemize}
\end{theorem}

Theorem~\ref{thm001} states that the proposed Bayesian approach successfully learns high-dimensional linear BNs. Note that the required assumptions ensure that $C_1$, $C_m$, $[\Sigma]_{j,j}$ remain constant for $(n,p,d_M)$ in Theorem~\ref{thm001}, which are involved with the graph recovery probability. Consequently, the proposed algorithm with appropriate hyper-parameters recovers the graph with high probability if $n= \Omega(d_M^2 \log p )$ for a sub-Gaussian linear BN and if $n = \Omega( d_M^2 p^{2/m})$ for a $4m$-th bounded-moment linear BN. The proof is built upon the related studies in \citet{ghoshal2018learning} and \citet{gan2019bayesian}, where the inverse covariance matrix-based algorithm and Bayesian regularized inverse covariance matrix estimation are considered, respectively. However, we combine these studies in a careful way to establish consistency in high-dimensional linear BN settings. The detailed proof can be found in Appendix.

Theorem~\ref{thm001} also shows that the proposed algorithm does not require commonly applied assumptions, such as the incoherence and faithfulness conditions (e.g., \citealp{kalisch2007estimating,uhler2013geometry,loh2014high,park2021learning,park2021learning2}). Instead, it requires closely related conditions on the hyper-parameters and thresholding parameter. In the following section, we demonstrate how these required conditions can be satisfied using certain special types of linear BNs.


\subsection{Illustration of the Assumptions and Hyper-Parameters}

\label{Secillustration}

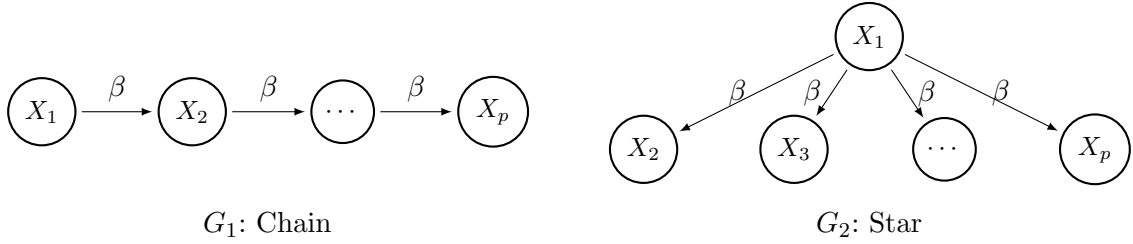
\begin{figure}
	\centering
	\begin{tikzpicture}[ -latex ,auto,
		state/.style={circle, draw=black, fill= white, thick, minimum size= 2mm},
		state2/.style={rectangle, draw=black, fill= white, thin, minimum size= 1.5mm},
		label/.style={thick, minimum size= 2mm}
		]
		
		\node[state] (X1)  at (-1,0)   {\small{$X_1$} }; 
		\node[state] (X2)  at (1,0)   {\small{$X_2$}}; 
		\node[state] (X3)  at (3,0)   {\small{$\cdots$}}; 
		\node[state] (X4)  at (5,0)   {\small{$X_p$}}; 
		\node[label] (G1) at (2,-1.5) {$G_1$: Chain};
		
		\path (X1) edge [shorten <= 2pt, shorten >= 2pt] node[above]  {$\beta$} (X2); 
		\path (X2) edge [shorten <= 2pt, shorten >= 2pt] node[above]  {$\beta$} (X3); 
		\path (X3) edge [shorten <= 2pt, shorten >= 2pt] node[above]  {$\beta$} (X4); 
		
		\node[state] (X1)  at (3+7,1.0)   {\small{$X_1$} }; 
		\node[state] (X2)  at (0+7,-0.5)   {\small{$X_2$}}; 
		\node[state] (X3)  at (2+7,-0.5)   {\small{$X_3$}}; 
		\node[state] (X4)  at (4+7,-0.5)   {\small{$\cdots$}}; 
		\node[state] (X5)  at (6+7,-0.5)   {\small{$X_p$}}; 
		\node[label] (G2) at (3+7,-1.5) {$G_2$: Star};
		
		\path (X1) edge [shorten <= 2pt, shorten >= 2pt] node[left]  {$\beta$} (X2); 
		\path (X1) edge [shorten <= 2pt, shorten >= 2pt] node[left]  {$\beta$} (X3); 
		\path (X1) edge [shorten <= 2pt, shorten >= 2pt] node[right]  {$\beta$} (X4); 
		\path (X1) edge [shorten <= 2pt, shorten >= 2pt] node[right]  {$\beta$} (X5);
	\end{tikzpicture}
	\caption{$p$-node chain and star graphs}
	\label{fig:illustration}
\end{figure}

This section confirms the validity of Assumptions~\ref{assu:eigen}, \ref{assu:condvar}, \ref{assu:min} and \ref{assu:spar}. Additionally, it provides appropriate hyper-parameters $v_0, v_1, \eta, \tau$ and the inclusion probability threshold $T$ when recovering chain and star linear BNs illustrated in Figure~\ref{fig:illustration}. These are popular sparse and dense models because the chain graph has a small maximum degree of the moralized graph $d_M = 2$, whereas the star graph has a large maximum degree $d_M = p-1$. More specifically, the considered chain and star linear BNs are as follow: For all $j \in \{1,2,...,p-1\}$,
$$
\text{Chain: } X_{j+1} = \beta X_{j} + \epsilon_{j+1}  \quad \text{ and } \quad
\text{Star: } X_{j+1} = \beta X_{1} + \epsilon_{j+1} \quad 
$$
where $X_1=\epsilon_1$ in both linear BNs and all the error variances are $\sigma^2$. Additionally, $\beta \in (-1, 1)$ is a small edge weight. 

In the chain linear BN, we can determine fixed bounds for $M_{\Gamma_{\max}}$, $M_{\Gamma_{\min}}$, and $M_\Sigma$, which are as follows:
\begin{align*}
	&M_{\Gamma_{\max}} = \left(\beta^4+2|\beta|^3+4\beta^2+2|\beta|+1\right)/\sigma^4, \
	M_{\Gamma_{\min}} = \left(\beta^4+2|\beta|^3+3\beta^2+2|\beta|+1\right)/\sigma^4, \\
	&M_\Sigma = \frac{(1-|\beta|^p)(1-|\beta|^{p+1})}{(1-|\beta|)(1-|\beta|^2)} \sigma^2 \leq \frac{1}{(1-|\beta])(1-|\beta|^2)} \sigma^2.
\end{align*}
Consequently, Assumption \ref{assu:spar} is satisfied if 
$M_1 >\frac{1}{(1-|\beta|)(1-|\beta|^2)} \sigma^2$, $M_2 > \frac{ \beta^4+2|\beta|^3+4\beta^2+2|\beta|+1 }{\sigma^4}$. Moreover, simple algebra yields that Assumptions~\ref{assu:eigen}, \ref{assu:condvar}, and \ref{assu:min} are satisfied if $k_1 \le \frac{\sigma^2}{(1+|\beta|)^2}$, $\tau_{\min} \le \frac{\beta^2}{\sigma^2}$, $\theta_{\min} \le \frac{|\beta|}{\sigma^2}$. 

However, in the star linear BN, we have the diverging values of $M_{\Gamma_{max}}$ and $M_\Sigma$ as the number of nodes increases:
\begin{align*}
	&M_{\Gamma_{max}} = \frac{  2(p-1)^2\beta^4+2(p-1)^2|\beta|^3+3(p-1)\beta^2+2(p-1)|\beta|+1 }{\sigma^4}, \\
	&M_{\Gamma_{min}} = \frac{ \left(2\beta^4+2|\beta|^3+3\beta^2+2|\beta|+1\right)}{ \sigma^4}, \ M_\Sigma = \max\left\{ (p-1)|\beta|+1, (p-1)\beta^2 + |\beta| +1\right\} \sigma^2.
\end{align*}
Consequently, Assumption \ref{assu:spar} rarely hold in large-scale linear BNs because of diverging $M_1$ and $M_2$. This heuristically supports that Assumption \ref{assu:spar} involves with the sparsity of the moralized graph. 

The success of the proposed approach depends on the existence of appropriate hyper-parameters. Hence, we now turn our attention to the existence of appropriate hyper-parameters $\nu_0$, $\nu_1$, $\tau$, and $T$ according to Theorem~\ref{thm001}. By setting suitable constants $C_1$, $C_2$, $C_3$, and $T$, we can easily find proper values for these hyper-parameters. A possible choice is as follows: For any $\epsilon_1>0$,
\begin{align*}
	&\nu_1 = \frac{p(1+\epsilon_1)}{nC_3},~\nu_0 = \frac{p}{nC_4},~\tau = \frac{nC_3}{2p},~\text{ and } T = \frac{\nu_0\eta}{\nu_1(1-\eta) + \nu_0\eta} ~\text{where}~\eta = 
	\frac{\nu_1^2}{\nu_1^2 + \nu_0^2\epsilon_1}.
\end{align*} 

In addition, for both chain and star linear BNs, we can set $C_1 = C_3/10$, $C_2 = d\theta_{\min} / (2M_{\Gamma_{max}})$, and
\begin{align*}
	C_3 = \frac{1}{2}\min\left\{
	\frac{1}{6M_{\Gamma_{max}}M_\Sigma},
	\frac{1}{6M_{\Gamma_{max}}^2M_\Sigma^3},
	\frac{k_1^2}{4M_{\Gamma_{max}}},
	\frac{\tau_{\min}}{4M_{\Gamma_{max}}},
	\frac{d\theta_{\min}}{2M_{\Gamma_{max}}},
	\frac{k_1^2p}{2\epsilon_1}\right\}.
\end{align*}

As discussed, the chain BN has fixed values for $M_{\Gamma_{\min}}$, $M_{\Gamma_{max}}$, and $M_\Sigma$, resulting in the fixed hyper-parameters regardless of the number of sample size and nodes. However, the star linear BN has diverging values for $M_{\Gamma_{max}}$ and $M_\Sigma$, leading to unacceptable value of hyper-parameters, such as $\nu_0$ and $\nu_1 $ diverges, $\tau \to 0$, and $T \to 1$, as $p$ increases in this setting.

So far, we have demonstrated that appropriate hyper-parameters exist for chain BNs regardless of the number of nodes or samples. However, finding appropriate hyper-parameters for large-scale star BN is difficult. Hence, this underscores the importance of Assumption~\ref{assu:spar} in achieving consistency, and highlights that the proposed method is well-suited for learning high-dimensional sparse linear BNs, while it may not be the optimal choice for recovering dense linear BNs.


Of course, the choice of required hyper-parameters depends on the true model quantities which are typically unknown in practice. Hence, in applications, we find ourselves in a similar position as for other graph learning algorithms (e.g., the PC, HGSM, BHLSM, and TD algorithms) where the output depends on test and tuning parameters. To select good hyper-parameter values for the BAGUS method, cross-validation can be applied as discussed in \citet{gan2019bayesian}. However, due to the heavy computational cost, we used fixed hyper-parameters, $v_0 = \sqrt{1/(100n)}, v_1 = 1, \tau = 0.0001,$ and $T = 0.5$, assuming unknown true model information in all our numerical experiments. The simulation results heuristically confirm that the proposed algorithm successfully recovers  sparse graphs with high probability. 

\section{Comparison to Frequentist Approaches}

\label{SecComparison}

This section compares Algorithm~\ref{OurAlgorithm} against other high-dimensional linear BN learning algorithms, such as BHLSM, TD, LISTEN, and graphical Lasso-based algorithms~\citep{loh2014high}, in terms of sample complexity and required assumptions. Regarding sample complexity, the $\ell_1$-regularized regression-based BHLSM algorithm can recover linear BNs with high probability if the sample sizes are $n = \Omega( d_M^{2} \log p)$ and $n = \Omega( d_M^2 p^{2/m} )$ for sub-Gaussian and $4m$-th bounded-moment error distributions, respectively. The best-subset-selection and $\ell_1$-regularized regression-combined TD algorithm can successfully learn a sub-Gaussian linear BN with high probability if the sample size is $n = \Omega(q^2 \log p)$, where $q$ is the predetermined upper bound of the maximum indegree. Finally, CLIME-based LISTEN algorithm can successfully learn a linear BN with high probability if the sample sizes are $n = \Omega(d_M^4 \log p)$ and $n = \Omega(d_M^4 p^{2/m})$ for sub-Gaussian and $4m$-th bounded-moment error distributions, respectively.

The proposed algorithm has a similar sample complexity and can learn a linear BN with high probability if the sample sizes are $n = \Omega( d_M^2 \log p )$ and $n = \Omega( d_M^2 p^{2/m} )$ with sub-Gaussian and $4m$-th bounded-moment linear BNs, respectively. The various simulation results in Section~\ref{SecNume} empirically support that the proposed and existing frequentist algorithms have similar performance when recovering directed edges.

In terms of required assumptions, most assumptions are similar. However, the BHLSM, TD, and graphical Lasso-based algorithms require the incoherence condition, which states that neighboring and non-neighboring nodes are not highly correlated. This condition may limit the applicability of these methods to certain sparse special graphs, such as the example of $1 \to 3, 4, 5$ and $2 \to 3, 4, 5$, where the neighboring node of $1$ is $(3,4,5)$, and the non-neighboring node of $1$ is $2$. However, it is structurally inevitable for non-neighboring node $2$ and neighborhood $(3,4,5)$ to be highly correlated. Hence, the proposed Bayesian approach can learn some models that other frequentist algorithms may fail to learn due to their different required assumptions, as long as appropriate hyper-parameters are applied.

\section{Numerical Experiments}

\label{SecNume}

This section presents empirical results to support our theoretical findings that Algorithm~\ref{OurAlgorithm} can consistently learn sub-Gaussian and $4m$-th bounded-moment linear BNs. Specifically, we consider three types of models: (i) Gaussian linear BNs, (ii) sub-Gaussian linear BNs with sequentially Uniform, Gaussian, and two-sided truncated Gaussian error distributions, and (iii) linear BNs with heavy-tailed error distributions, where student t-distributions with 10 degrees of freedom are applied in both low- and high-dimensional settings. Additionally, this section compares the performance of Algorithm~\ref{OurAlgorithm} with that of LISTEN~\citep{ghoshal2018learning}, BHLSM~\citep{park2021learning2}, and TD~\citep{chen2019causal} algorithms in terms of edge recovery.

The hyper-parameters for the proposed algorithm were set to $v_0 = \sqrt{1/(100n)}, v_1 = 1, \tau = 0.0001, T = 0.5$, under the assumption of unknown true model information. These parameters could have been selected through cross-validation, as suggested in \citet{gan2019bayesian}. However, to facilitate faster implementation, we applied fixed values while respecting a small spike ($v_0$) and tuning ($\tau$) parameters, as recommended in Section~\ref{SecAlgo}.

For our simulation settings, we slightly modified the LISTEN and TD algorithms to improve their performance and stability. Specifically, the modified LISTEN algorithm applied CLIME to estimate each element of the ordering, instead of using it only to estimate the last element and then applying a matrix update approach. This was necessary because the original LISTEN algorithm often failed due to accumulated error in matrix updates. Additionally, we set the regularized regression parameter to $\sqrt{\frac{\log p}{n}}$ and the hard threshold parameter to half the minimum value of true edge weights, $\min_{(j,k) \in E} (|\beta_{j,k}|/2)$. The modified TD algorithm used $\ell_1$-regularized regression for parent estimation, similar to the BHLSM algorithm. The regularized parameter was set to $3\sqrt{\frac{\log p}{n}}$ to achieve higher accuracy in our settings. Finally, we set $q$ to the maximum degree of the true moralized graph, $d_M$.

The proposed algorithm and comparison algorithms were evaluated by measuring the average hamming distance between the estimated and true DAGs while varying sample sizes. The hamming distance was calculated as the number of edges that differ between the two graphs; hence, a smaller value is better.

\subsection{Gaussian Linear BNs}

\label{SecNume001}

We conducted simulations using $100$ realizations of $p$-node Gaussian linear BNs with randomly generated underlying DAG structures for node size $p \in \{25, 50, 100, 150, 200\}$ while respecting the maximum degree constraint $d_M \in \{3, 5, 8\}$, as done by \citet{ghoshal2017learning,park2021learning2}. Specifically, Erd\"{o}s and R\'{e}nyi graphs were considered with edge probability $q = \min  (1, 3 d_M/ p)$. If the maximum degree of the moralized graph was greater than pre-determined $d_M$, we generated the graph again with an updated edge probability of $q - 0.001$ until the maximum degree condition was satisfied. We generated non-zero edge weights uniformly at random in the range $\beta_{k,j} \in (-1.0, -0.5) \cup (0.5, 1.0)$. Finally, we set all noise variances to $\sigma_j^2 = 2$.

\begin{figure*}[!t] \vspace{-0mm}
	\centering \hspace{-3mm}
	\begin{subfigure}[!htb]{.30\textwidth}
		\includegraphics[width=\textwidth, trim={0.75cm 0cm 0cm 0cm},clip]{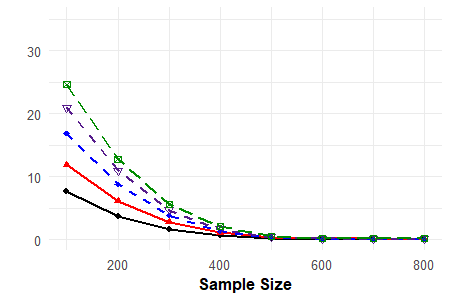}
		\caption{$d_M = 3$}
	\end{subfigure} \hspace{-3mm}
	\begin{subfigure}[!htb]{.30\textwidth}
		\includegraphics[width=\textwidth, trim={0.75cm 0cm 0cm 0cm},clip]{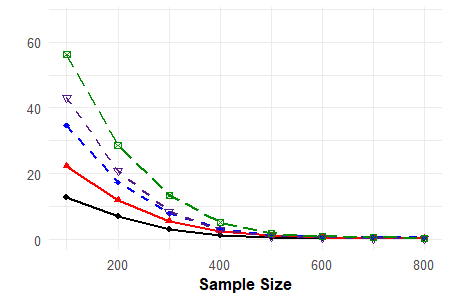}
		\caption{$d_M = 5$}
	\end{subfigure}	\hspace{-3mm}
	\begin{subfigure}[!htb]{.30\textwidth}
		\includegraphics[width=\textwidth, trim={0.75cm 0cm 0cm 0cm},clip]{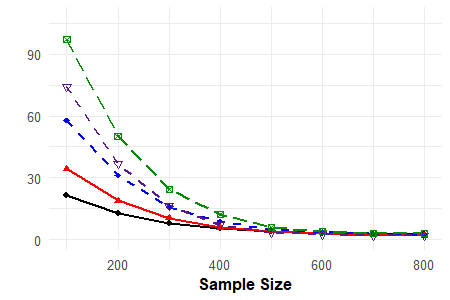}
		\caption{$d_M = 8$}
	\end{subfigure}	\hspace{-3mm}
	\begin{subfigure}[!htb]{.10\textwidth}
		\includegraphics[width=\textwidth, trim={13.75cm 0.5cm 0.4cm 1.0cm},clip]{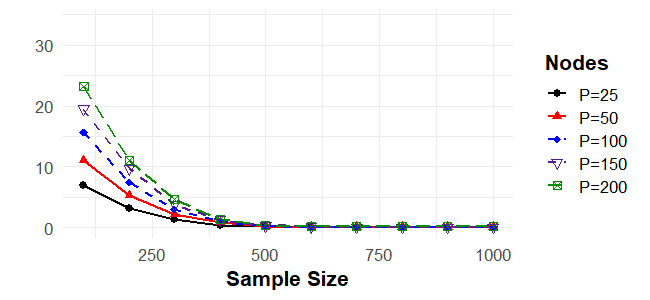}
	\end{subfigure}
	\centering \hspace{-3mm}
	\begin{subfigure}[!htb]{.30\textwidth}
		\includegraphics[width=\textwidth, trim={0.75cm 0cm 0cm 0cm},clip]{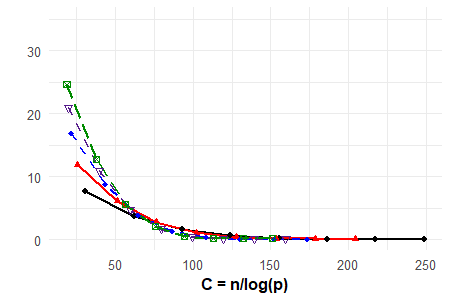}
		\caption{$d_M = 3$}
	\end{subfigure} \hspace{-3mm}
	\begin{subfigure}[!htb]{.30\textwidth}
		\includegraphics[width=\textwidth, trim={0.75cm 0cm 0cm 0cm},clip]{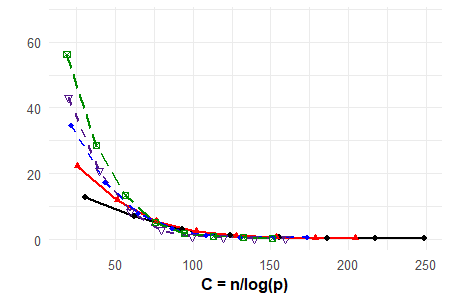}
		\caption{$d_M = 5$}
	\end{subfigure}	\hspace{-3mm}
	\begin{subfigure}[!htb]{.30\textwidth}
		\includegraphics[width=\textwidth, trim={0.75cm 0cm 0cm 0cm},clip]{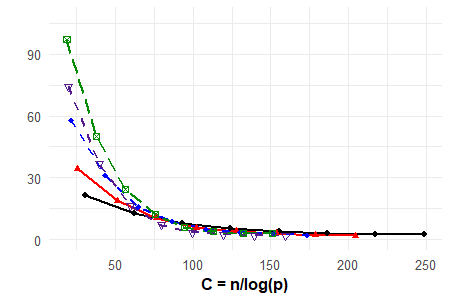}
		\caption{$d_M = 8$}
	\end{subfigure}	\hspace{-3mm}
	\begin{subfigure}[!htb]{.10\textwidth}
		\includegraphics[width=\textwidth, trim={13.75cm 0.5cm 0.4cm 1.0cm},clip]{figures/Legend002.png}
	\end{subfigure}
	\caption{
		Average hamming distances for Gaussian linear BNs with maximum degree $d_M \in \{3,5,8\}$ plotted against the sample size $n$ (top) and the re-scaled sample size $C=n/(\log p)$ (bottom).
	}
	\label{fig:result001}
\end{figure*}

Figures~\ref{fig:result001} (a) - (c) show the average hamming distance of Algorithm~\ref{OurAlgorithm} by varying sample size $n \in \{100, 200, ..., 800\}$. Figures~\ref{fig:result001} (d) - (f) show the hamming distance against re-scaled sample size $C = n / \log p$. As seen in Figure~\ref{fig:result001}, the proposed algorithm recovers the true directed edges better as the sample size increases and the hamming distance converges to 0. Additionally, the empirical curves for different numbers of nodes align more closely with the re-scaled sample size on the horizontal axis. This supports the main result in Theorem~\ref{thm001} that number of samples $n$ required for successful graph recovery scales logarithmically with number of nodes $p$ in Gaussian linear BNs.

Figure~\ref{fig:result001} also reveals that Algorithm~\ref{OurAlgorithm} requires fewer samples to recover a sparse graph. Specifically, the average hamming distances for 50-node graphs with $d_M = 3$ and $8$ are approximately $5$ and $20$, respectively, when $n = 200$. Similar phenomena are shown for all other considered number of samples and nodes. Hence, the simulation results confirm our theoretical findings that the sample complexity of the proposed algorithm relies on the maximum degree.

\begin{figure*}[!t] \vspace{-0mm}
	\centering \hspace{-3mm}
	\begin{subfigure}[!htb]{.30\textwidth}
		\includegraphics[width=\textwidth, trim={0.75cm 0cm 0cm 0cm},clip]{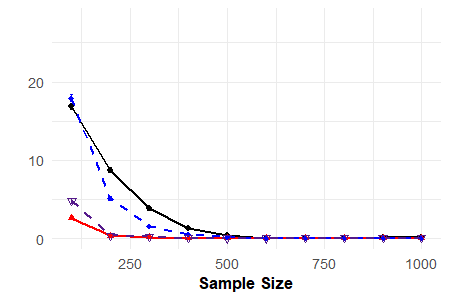}
		\caption{$p = 100$, $d_M = 3$}
	\end{subfigure} \hspace{-3mm}
	\begin{subfigure}[!htb]{.30\textwidth}
		\includegraphics[width=\textwidth, trim={0.75cm 0cm 0cm 0cm},clip]{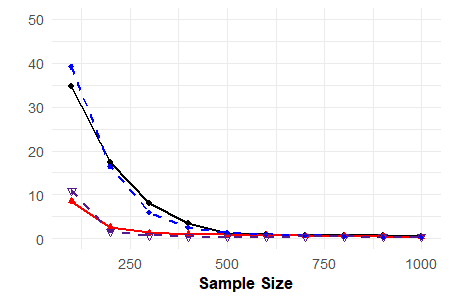}
		\caption{$p = 100$, $d_M = 5$}
	\end{subfigure} \hspace{-3mm}
	\begin{subfigure}[!htb]{.30\textwidth}
		\includegraphics[width=\textwidth, trim={0.75cm 0cm 0cm 0cm},clip]{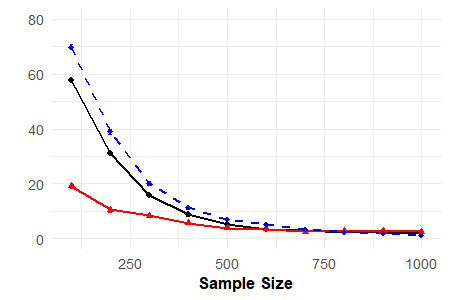}
		\caption{$p = 100$, $d_M = 8$}
	\end{subfigure}	\hspace{-3mm}
	\begin{subfigure}[!htb]{.10\textwidth}
		\includegraphics[width=\textwidth, trim={13.25cm 0.5cm 0.4cm 1.0cm},clip]{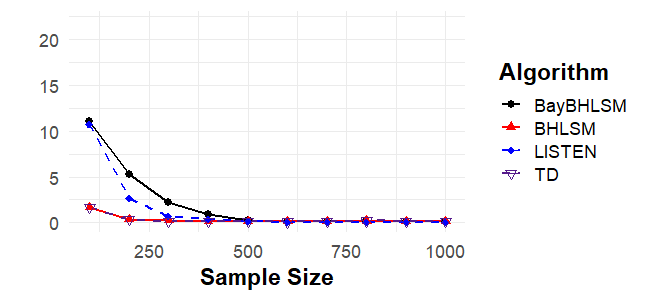}
	\end{subfigure}
	
	\begin{subfigure}[!htb]{.30\textwidth}
		\includegraphics[width=\textwidth, trim={0.75cm 0cm 0cm 0cm},clip]{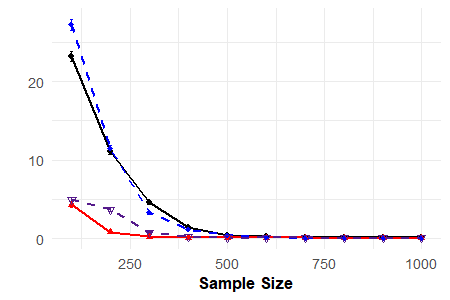}
		\caption{$p = 200$, $d_M = 3$}
	\end{subfigure} \hspace{-3mm}
	\begin{subfigure}[!htb]{.30\textwidth}
		\includegraphics[width=\textwidth, trim={0.75cm 0cm 0cm 0cm},clip]{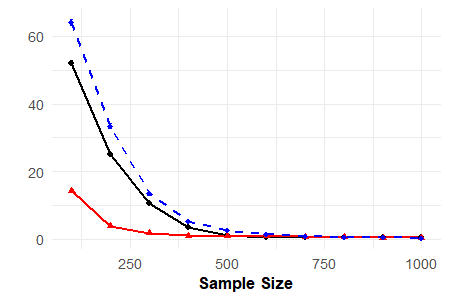}
		\caption{$p = 200$, $d_M = 5$}
	\end{subfigure} \hspace{-3mm}
	\begin{subfigure}[!htb]{.30\textwidth}
		\includegraphics[width=\textwidth, trim={0.75cm 0cm 0cm 0cm},clip]{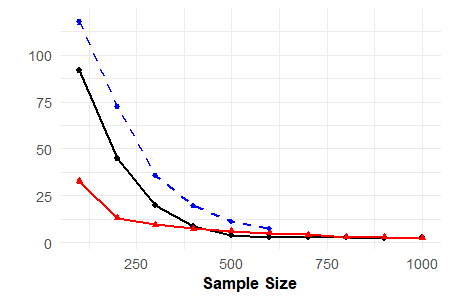}
		\caption{$p = 200$, $d_M = 8$}
	\end{subfigure}	\hspace{-3mm}
	\begin{subfigure}[!htb]{.10\textwidth}
		\includegraphics[width=\textwidth, trim={13.25cm 0.5cm 0.4cm 1.0cm},clip]{figures/Legend001.png}
	\end{subfigure}
	\caption{
		Comparison of the proposed algorithm (BayBHLSM) against the BHLSM, LISTEN, and TD algorithms in terms of average hamming distance for learning Gaussian linear BNs.
	}
	\label{fig:result002}
\end{figure*}

Figure~\ref{fig:result002} evaluates the proposed algorithm (BayBHLSM) and the frequentist alternatives, BHLSM, LISTEN, and TD algorithms, in terms of recovering DAGs with $p \in \{100, 200\}$ and $d_M \in \{3,5,8\}$ by varying $n \in \{100, 200, ..., 1000\}$. As shown in Figure~\ref{fig:result002}, the proposed algorithm generally performs as accurately as the comparison algorithms. This reflects that the proposed and comparison algorithms have similar sample complexities. Additionally, the Lasso-based BHLSM and TD algorithms perform better when sample size is small due to the choice of the regularization parameter, not because of the superiority of the frequentist approaches. However, we can also see that the proposed algorithm performs better when the sample size is sufficiently large.

A major drawback of the proposed algorithm is its computational cost. For example, the proposed algorithm takes about 5 minutes and 1.5 hours, on average, when learning 100- and 200-node graphs with $d_M = 8$ given a sample size of $n = 100$, respectively. This is consistent with the computational complexity of the proposed algorithm discussed in Section~\ref{SecAlgo}, where it is sensitive to the number of nodes ($O(p^4)$). Nevertheless, it is faster than the best-subset-selection-based TD algorithm with computational complexity of at least $O(p^{q+1}q^3)$. For example, the TD algorithm takes more than 12 hours to learn a 100-node graph with $d_M = 8$. Hence, we do not present the results of the TD algorithm for large-scale graphs with $p = 200$ and $d_M \geq 5$ due to the huge run time.

\subsection{Sub-Gaussian Linear BNs with Different Error Distributions}

\label{SecNume002}

\begin{figure*}[!t] \vspace{-0mm}
	\centering \hspace{-3mm}
	\begin{subfigure}[!htb]{.30\textwidth}
		\includegraphics[width=\textwidth, trim={0.75cm 0cm 0cm 0cm},clip]{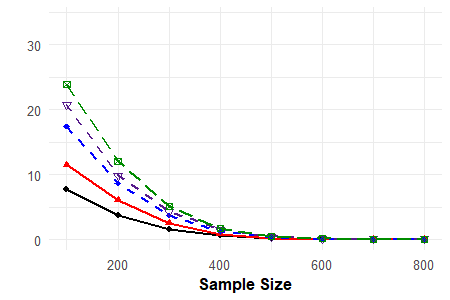}
		\caption{$d_M = 3$}
	\end{subfigure} \hspace{-3mm}
	\begin{subfigure}[!htb]{.30\textwidth}
		\includegraphics[width=\textwidth, trim={0.75cm 0cm 0cm 0cm},clip]{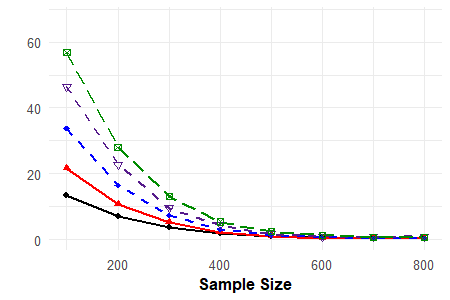}
		\caption{$d_M = 5$}
	\end{subfigure}	\hspace{-3mm}
	\begin{subfigure}[!htb]{.30\textwidth}
		\includegraphics[width=\textwidth, trim={0.75cm 0cm 0cm 0cm},clip]{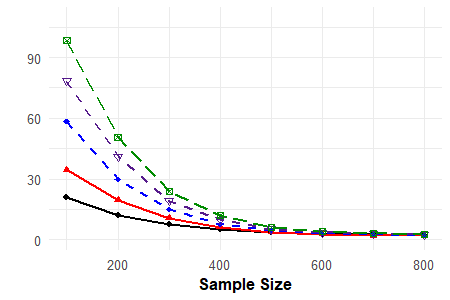}
		\caption{$d_M = 8$}
	\end{subfigure}	\hspace{-3mm}
	\begin{subfigure}[!htb]{.10\textwidth}
		\includegraphics[width=\textwidth, trim={13.75cm 0.5cm 0.4cm 1.0cm},clip]{figures/Legend002.png}
	\end{subfigure}
	\centering \hspace{-3mm}
	\begin{subfigure}[!htb]{.30\textwidth}
		\includegraphics[width=\textwidth, trim={0.75cm 0cm 0cm 0cm},clip]{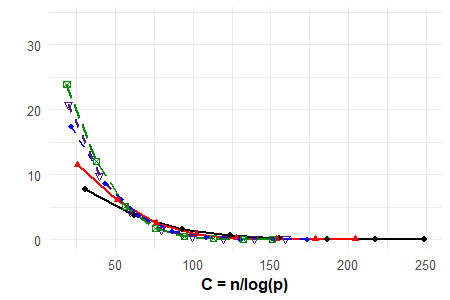}
		\caption{$d_M = 3$}
	\end{subfigure} \hspace{-3mm}
	\begin{subfigure}[!htb]{.30\textwidth}
		\includegraphics[width=\textwidth, trim={0.75cm 0cm 0cm 0cm},clip]{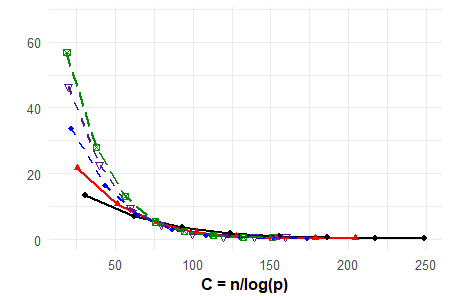}
		\caption{$d_M = 5$}
	\end{subfigure}	\hspace{-3mm}
	\begin{subfigure}[!htb]{.30\textwidth}
		\includegraphics[width=\textwidth, trim={0.75cm 0cm 0cm 0cm},clip]{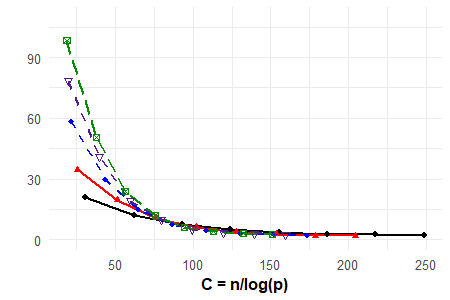}
		\caption{$d_M = 8$}
	\end{subfigure}	\hspace{-3mm}
	\begin{subfigure}[!htb]{.10\textwidth}
		\includegraphics[width=\textwidth, trim={13.75cm 0.5cm 0.4cm 1.0cm},clip]{figures/Legend002.png}
	\end{subfigure}
	\caption{
		Average hamming distances for Sub-Gaussian linear BNs with maximum degree $d_M \in \{3,5,8\}$ plotted against the sample size $n$ (top) and the re-scaled sample size $C=n/(\log p)$ (bottom).
	}
	\label{fig:result003}
\end{figure*}

\begin{figure*}[!t] \vspace{-0mm}
	\centering \hspace{-3mm}
	\begin{subfigure}[!htb]{.30\textwidth}
		\includegraphics[width=\textwidth, trim={0.75cm 0cm 0cm 0cm},clip]{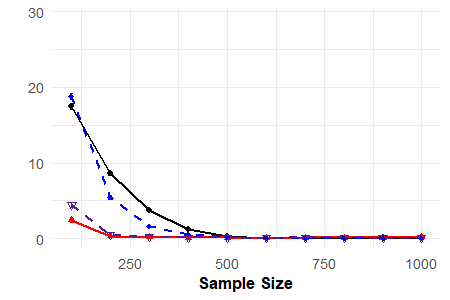}
		\caption{$p = 100$, $d_M = 3$}
	\end{subfigure} \hspace{-3mm}
	\begin{subfigure}[!htb]{.30\textwidth}
		\includegraphics[width=\textwidth, trim={0.75cm 0cm 0cm 0cm},clip]{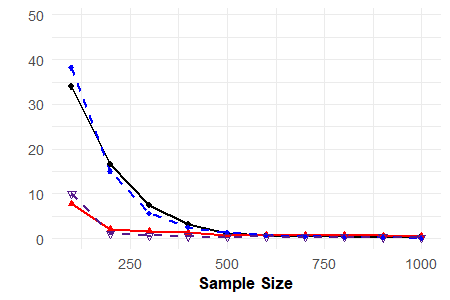}
		\caption{$p = 100$, $d_M = 5$}
	\end{subfigure} \hspace{-3mm}
	\begin{subfigure}[!htb]{.30\textwidth}
		\includegraphics[width=\textwidth, trim={0.75cm 0cm 0cm 0cm},clip]{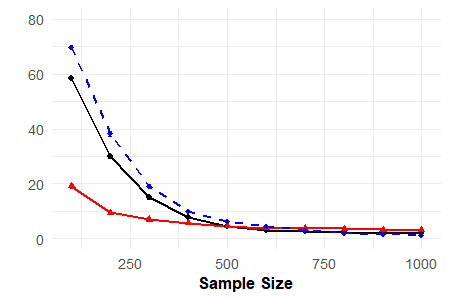}
		\caption{$p = 100$, $d_M = 8$}
	\end{subfigure}	\hspace{-3mm}
	\begin{subfigure}[!htb]{.10\textwidth}
		\includegraphics[width=\textwidth, trim={13.25cm 0.5cm 0.4cm 1.0cm},clip]{figures/Legend001.png}
	\end{subfigure}
	
	\centering \hspace{-3mm}
	\begin{subfigure}[!htb]{.30\textwidth}
		\includegraphics[width=\textwidth, trim={0.75cm 0cm 0cm 0cm},clip]{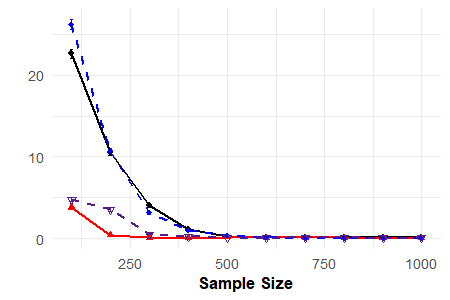}
		\caption{$p = 200$, $d_M = 3$}
	\end{subfigure} \hspace{-3mm}
	\begin{subfigure}[!htb]{.30\textwidth}
		\includegraphics[width=\textwidth, trim={0.75cm 0cm 0cm 0cm},clip]{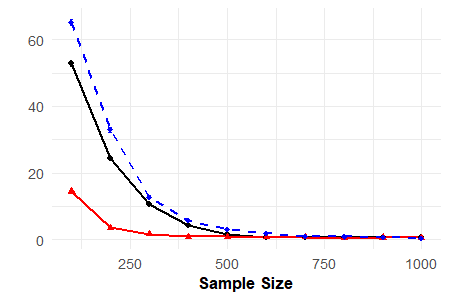}
		\caption{$p = 200$, $d_M = 5$}
	\end{subfigure} \hspace{-3mm}
	\begin{subfigure}[!htb]{.30\textwidth}
		\includegraphics[width=\textwidth, trim={0.75cm 0cm 0cm 0cm},clip]{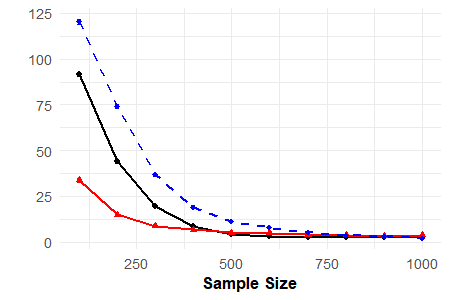}
		\caption{$p = 200$, $d_M = 8$}
	\end{subfigure}	\hspace{-3mm}
	\begin{subfigure}[!htb]{.10\textwidth}
		\includegraphics[width=\textwidth, trim={13.25cm 0.5cm 0.4cm 1.0cm},clip]{figures/Legend001.png}
	\end{subfigure}
	
	\caption{
		Comparison of the proposed algorithm (BayBHLSM) against the BHLSM, LISTEN, and TD algorithms in terms of average hamming distance for learning Sub-Gaussian linear BNs.
	}
	\label{fig:result004}
\end{figure*}

This section considers 
sub-Gaussian linear BNs where heterogeneous error variances and non-Gaussian error distributions are allowed. Hence, we generated 100 sets of samples following the procedure specified in Section~\ref{SecNume001}, except that error distributions were sequentially Uniform ($U(-2.5, 2.5)$), Gaussian ($N(0, 2)$), and two-sided truncated Gaussian ($N(0, 10)$ within the interval $(-2.5, 2.5)$). Figures \ref{fig:result003} and \ref{fig:result004} evaluate the proposed algorithm and the comparison methods in terms of the hamming distance by varying the sample size.

The simulation results in Figures~\ref{fig:result003} and \ref{fig:result004} are analogous to the results for Gaussian linear BNs with the same error variances presented in Section~\ref{SecNume001}. Specifically, Figure~\ref{fig:result003} empirically supports the theoretical result that the proposed algorithm requires sample size $n$ depending on maximum degree $d_M$ and $\log p$ for successful graph recovery. However, Figure~\ref{fig:result003} also shows that the hamming distance does not reach zero when $d_M = 8$. This reflects the fact that $\tau_{\min}$ in the minimum gap condition can be small when error variances are different. Hence, the proposed algorithm requires a larger number of samples to recover a graph. Nonetheless, Figure~\ref{fig:result003} confirms that Algorithm~\ref{OurAlgorithm} can consistently learn high-dimensional sparse linear BNs, even when error distributions are non-Gaussian, and error variances are different. Figure~\ref{fig:result004} shows that the proposed algorithm at our settings recovers the graph as accurately as the frequentist algorithms in terms of hamming distance.

\subsection{T Linear BNs}

This section considers linear BNs with heavy-tailed error distributions. 

\begin{figure*}[!t] \vspace{-0mm}
	\centering \hspace{-3mm}
	\begin{subfigure}[!htb]{.30\textwidth}
		\includegraphics[width=\textwidth, trim={0.75cm 0cm 0cm 0cm},clip]{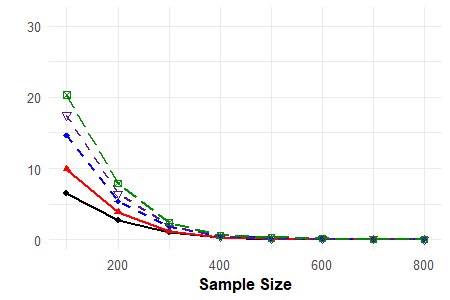}
		\caption{$d_M = 3$}
	\end{subfigure} \hspace{-3mm}
	\begin{subfigure}[!htb]{.30\textwidth}
		\includegraphics[width=\textwidth, trim={0.75cm 0cm 0cm 0cm},clip]{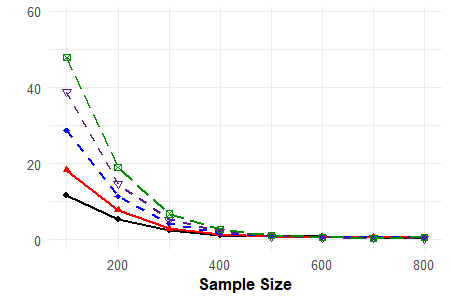}
		\caption{$d_M = 5$}
	\end{subfigure}	\hspace{-3mm}
	\begin{subfigure}[!htb]{.30\textwidth}
		\includegraphics[width=\textwidth, trim={0.75cm 0cm 0cm 0cm},clip]{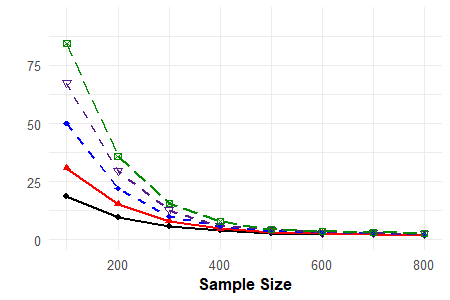}
		\caption{$d_M = 8$}
	\end{subfigure}	\hspace{-3mm}
	\begin{subfigure}[!htb]{.10\textwidth}
		\includegraphics[width=\textwidth, trim={13.75cm 0.5cm 0.4cm 1.0cm},clip]{figures/Legend002.png}
	\end{subfigure}
	\centering \hspace{-3mm}
	\begin{subfigure}[!htb]{.30\textwidth}
		\includegraphics[width=\textwidth, trim={0.75cm 0cm 0cm 0cm},clip]{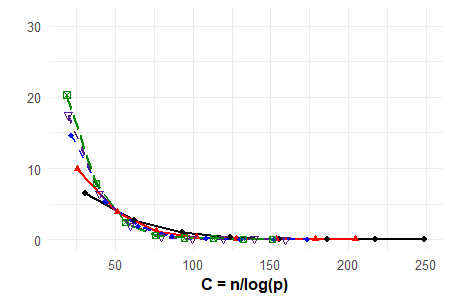}
		\caption{$d_M = 3$}
	\end{subfigure} \hspace{-3mm}
	\begin{subfigure}[!htb]{.30\textwidth}
		\includegraphics[width=\textwidth, trim={0.75cm 0cm 0cm 0cm},clip]{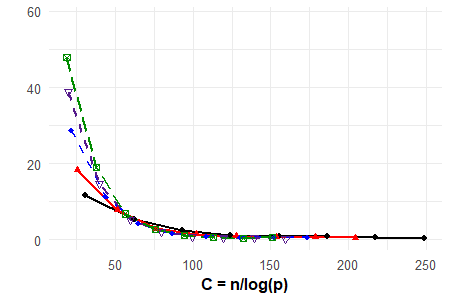}
		\caption{$d_M = 5$}
	\end{subfigure}	\hspace{-3mm}
	\begin{subfigure}[!htb]{.30\textwidth}
		\includegraphics[width=\textwidth, trim={0.75cm 0cm 0cm 0cm},clip]{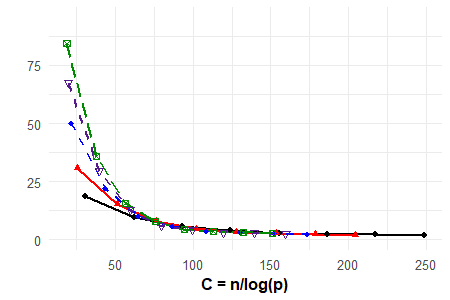}
		\caption{$d_M = 8$}
	\end{subfigure}	\hspace{-3mm}
	\begin{subfigure}[!htb]{.10\textwidth}
		\includegraphics[width=\textwidth, trim={13.75cm 0.5cm 0.4cm 1.0cm},clip]{figures/Legend002.png}
	\end{subfigure}
	\caption{
		Average hamming distances for T linear BNs with maximum degree $d_M \in \{3,5,8\}$ plotted against the sample size $n$ (top) and the re-scaled sample size $C=n/(\log p)$ (bottom)
	}
	\label{fig:result005}
\end{figure*}

\begin{figure*}[!t] \vspace{-0mm}
	\centering \hspace{-3mm}
	\begin{subfigure}[!htb]{.30\textwidth}
		\includegraphics[width=\textwidth, trim={0.75cm 0cm 0cm 0cm},clip]{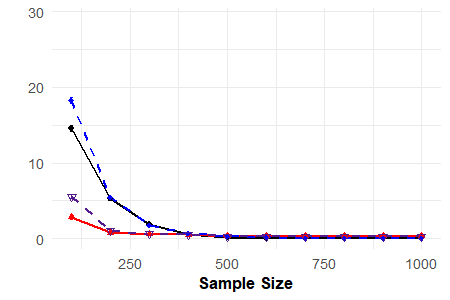}
		\caption{$p = 100$, $d_M = 3$}
	\end{subfigure} \hspace{-3mm}
	\begin{subfigure}[!htb]{.30\textwidth}
		\includegraphics[width=\textwidth, trim={0.75cm 0cm 0cm 0cm},clip]{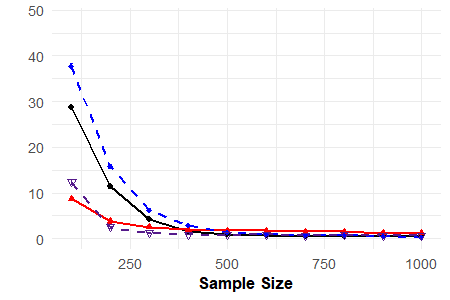}
		\caption{$p = 100$, $d_M = 5$}
	\end{subfigure} \hspace{-3mm}
	\begin{subfigure}[!htb]{.30\textwidth}
		\includegraphics[width=\textwidth, trim={0.75cm 0cm 0cm 0cm},clip]{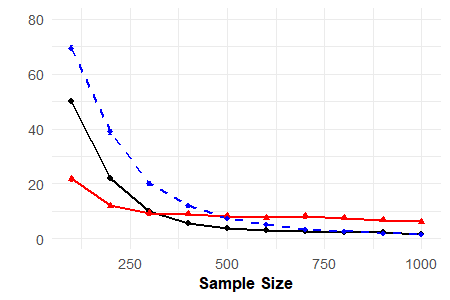}
		\caption{$p = 100$, $d_M = 8$}
	\end{subfigure}	\hspace{-3mm}
	\begin{subfigure}[!htb]{.10\textwidth}
		\includegraphics[width=\textwidth, trim={13.25cm 0.5cm 0.4cm 1.0cm},clip]{figures/Legend001.png}
	\end{subfigure}
	
	\centering \hspace{-3mm}
	\begin{subfigure}[!htb]{.30\textwidth}
		\includegraphics[width=\textwidth, trim={0.75cm 0cm 0cm 0cm},clip]{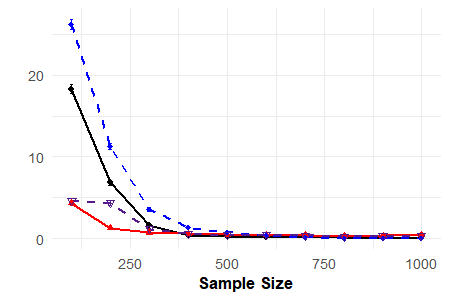}
		\caption{$p = 200$, $d_M = 3$}
	\end{subfigure} \hspace{-3mm}
	\begin{subfigure}[!htb]{.30\textwidth}
		\includegraphics[width=\textwidth, trim={0.75cm 0cm 0cm 0cm},clip]{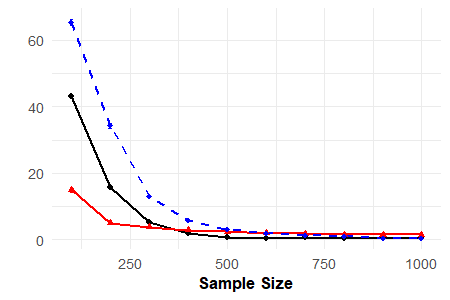}
		\caption{$p = 200$, $d_M = 5$}
	\end{subfigure} \hspace{-3mm}
	\begin{subfigure}[!htb]{.30\textwidth}
		\includegraphics[width=\textwidth, trim={0.75cm 0cm 0cm 0cm},clip]{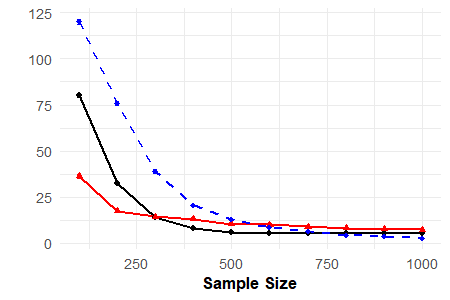}
		\caption{$p = 200$, $d_M = 8$}
	\end{subfigure}	\hspace{-3mm}
	\begin{subfigure}[!htb]{.10\textwidth}
		\includegraphics[width=\textwidth, trim={13.25cm 0.5cm 0.4cm 1.0cm},clip]{figures/Legend001.png}
	\end{subfigure}
	
	\caption{
		Comparison of the proposed algorithm (BayBHLSM) against the BHLSM, LISTEN, and TD algorithms in terms of average hamming distance for learning T linear BNs.
	}
	\label{fig:result006}
\end{figure*}

Hence, we generated 100 sets of samples under the procedure specified in Sections~\ref{SecNume001}, except that error distributions were student t-distributions with 10 degrees of freedom. The performance of the proposed algorithm and the comparison algorithms in terms of the hamming distance is presented in Figures~\ref{fig:result005} and \ref{fig:result006}.

The results shown in Figures~\ref{fig:result005} and \ref{fig:result006} are similar to the previous simulation results for (sub-)Gaussian linear BNs in Sections~\ref{SecNume001} and \ref{SecNume002}. More specifically, Figure~\ref{fig:result005} confirms that the proposed algorithm can consistently learn high-dimensional sparse linear BNs with heavy-tailed error distributions. Additionally, Figure~\ref{fig:result006} shows that the proposed algorithm in our settings recovers the graph as accurately as the frequentist algorithms in terms of hamming distance.

\section{Real Data}

\label{SecReal}

The proposed algorithm was applied to real-world order data from an online shopping mall, which is available at \url{https://www.kaggle.com/datasets/mervemenekse/ecommerce-dataset}. 
The original dataset contains the order history of $38997$ customers and $42$ products. It includes information such as customer ID, gender, product category, and product name. In this analysis, we focused on finding the relationships between product order amounts using only the number of orders for each product to highlight the advantages of the proposed algorithm. Hence, the data considered in this analysis consists of $n=38997$ observations and $p=42$ variables, where the $i$-th row and $j$-th column represent the number of orders for the $j$-th product by the $i$-th customer at the online mall.

We begin by describing some characteristics of the orders by category. All products on sale at the online mall are divided into four categories:  `Auto $\&$ Accessories', `Fashion', `Electronic', and `Home $\&$ Furniture'. Specifically, each category contains a similar numbers of products, with 9, 11, 12, and 10 products in the `Auto $\&$ Accessories', `Fashion', `Electronic', and `Home $\&$ Furniture' categories, respectively. However, most orders are concentrated in two categories, `Fashion' and `Home $\&$ Furniture', with $57.28\%$ of orders for fashion-related products and $32.90\%$ of orders for home-related products. Furthermore, $91.76\%$ of customers made six or fewer purchases, which suggests that most nodes may not be connected by edges.

\begin{figure*}[!t] \vspace{-0mm}
	\centering \hspace{-3mm}
	\begin{subfigure}[!htb]{.40\textwidth}
		\includegraphics[width=\textwidth, trim={0cm 0cm 5cm 0cm},clip]{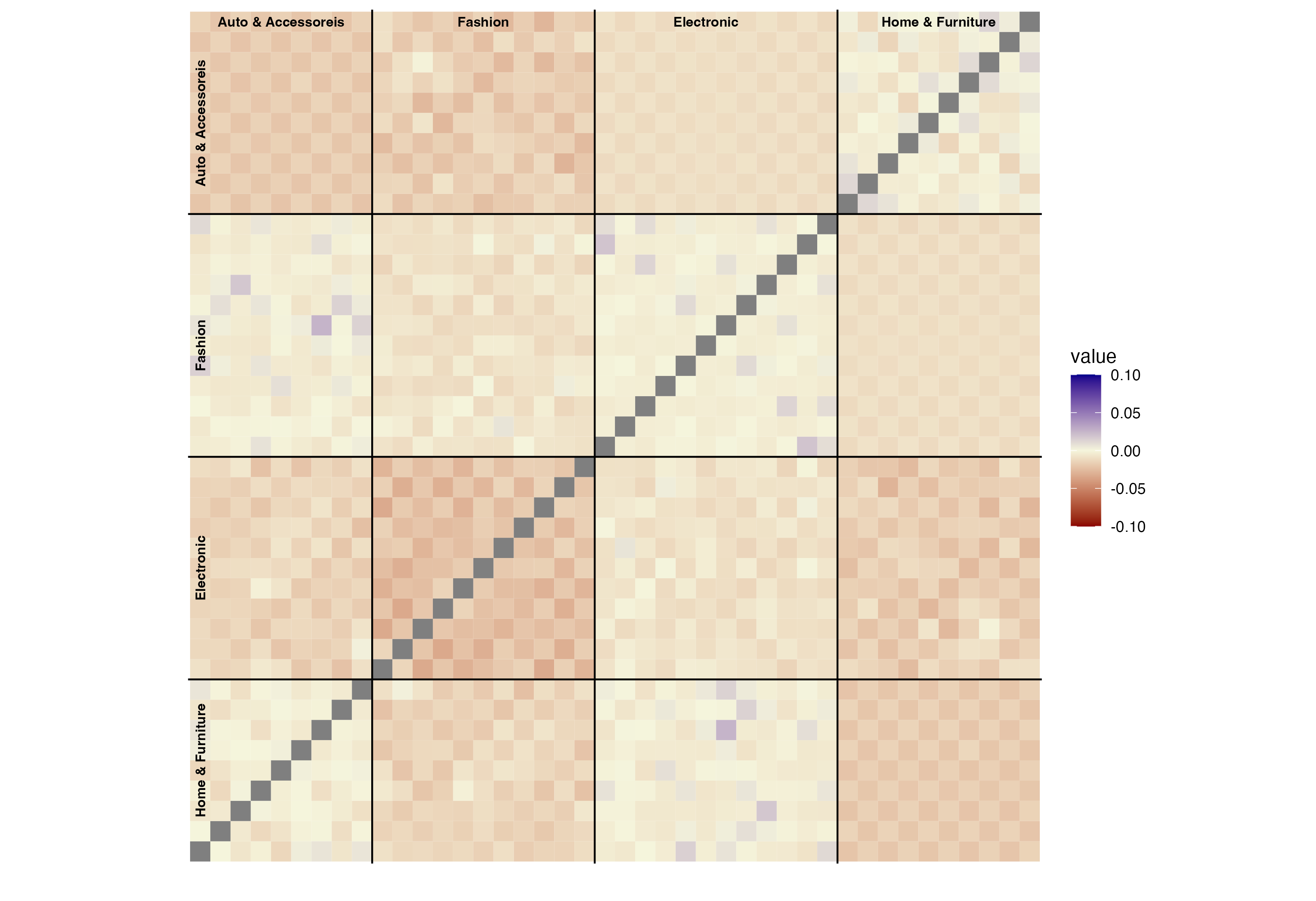}
		\caption{Correlation Plot}
	\end{subfigure}
	\begin{subfigure}[!htb]{.40\textwidth}
		\includegraphics[width=\textwidth, trim={0cm 0cm 5cm 0cm},clip]{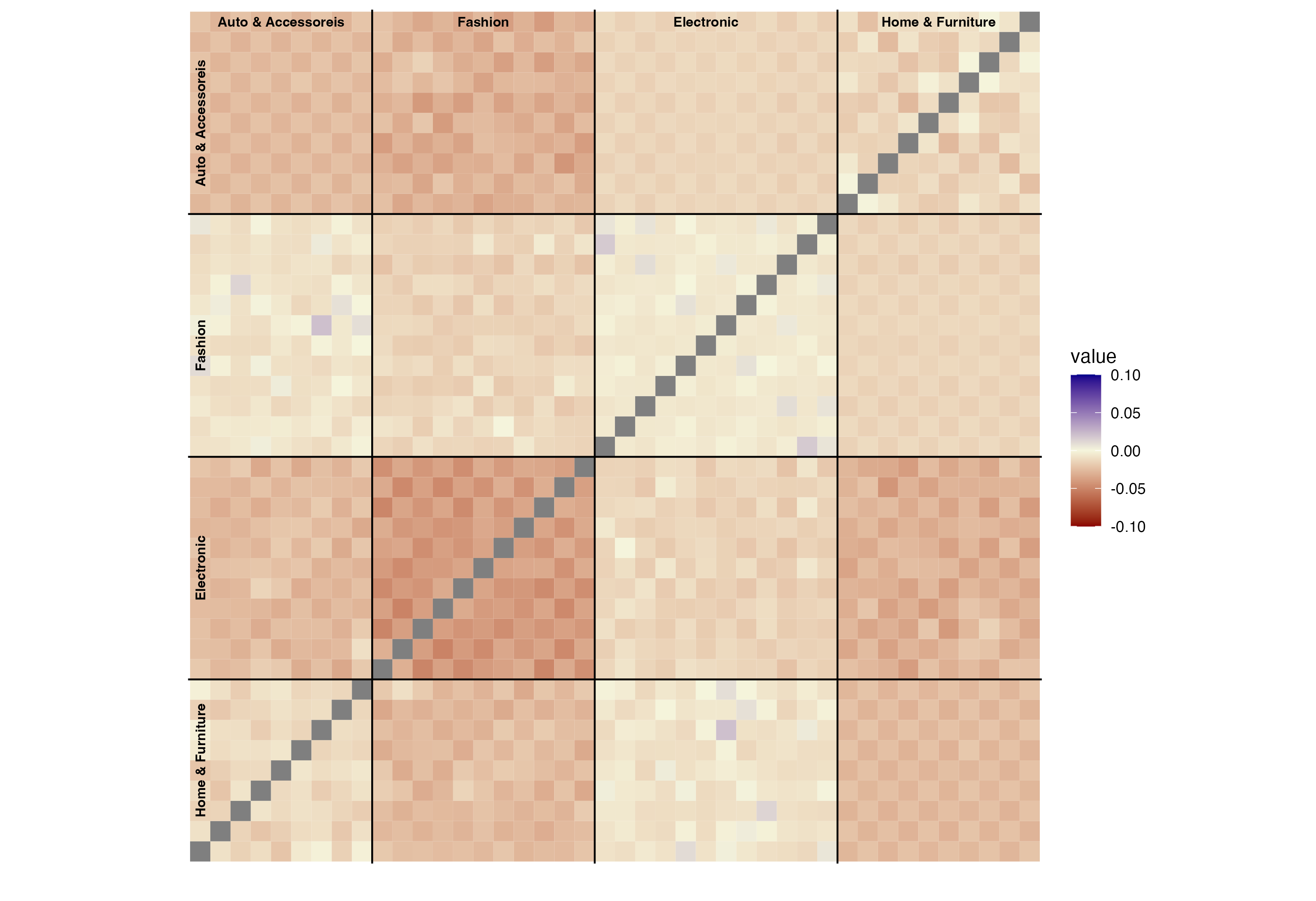}
		\caption{Partial Correlation Plot}
	\end{subfigure}	
	\begin{subfigure}[!htb]{.15\textwidth}
		\includegraphics[width=\textwidth, trim={21cm 7cm 0.1cm 7cm},clip]{figures/pp01.png}
	\end{subfigure}	
	\caption{
		Correlation and partial correlation plots for order amounts by products.
	}
	\label{fig:corr}
\end{figure*}

In Figure \ref{fig:corr}, the correlation and partial correlation plots are depicted, revealing the near absence of (conditional) correlation among all 42 variables. Notably, the partial correlation plot in Figure~\ref{fig:corr} (b) demonstrates that fashion-related products exhibit modest conditional correlations when accounting for other variables. This makes sense because $10.04\%$ of customers bought multiple fashion-related products, which is a higher percentage than for other categories. The percentages of customers who bought multiple products belonging to the same category are $8.45\%$, $7.88\%$, and $2.33\%$ in `Auto \& Accessories', `Electronic', and `Home \& Furniture' categories, respectively. Hence, one can expect a sparse underlying graph structure although some nodes are highly connected.


\begin{figure*}[!t] \vspace{-0mm}
	\centering \hspace{-3mm}
	\begin{subfigure}[!htb]{.70\textwidth}
		\includegraphics[width=\textwidth, trim={0cm 0cm 0cm 0cm},clip]{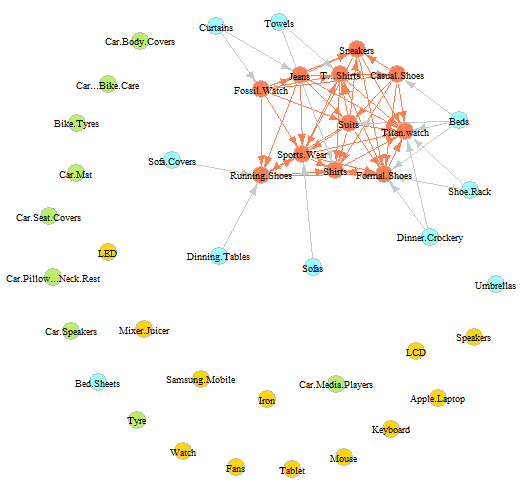}
	\end{subfigure}
	\caption{
		Estimated directed acyclic graph via the proposed algorithm.
		The color of each node indicates the category to which it belongs.
		`Auto $\&$ Accessories': green.
		`Fashion': orange.
		`Electronic': yellow.
		`Home $\&$ Furniture': sky blue.
	}
	\label{fig:graph}
\end{figure*}

Figure~\ref{fig:graph} shows the directed graph estimated by the proposed algorithm. Since the true model information is unknown, the hyper-parameters of the proposed algorithm are set as $\nu_0 = \sqrt{1/(100n)}$, $\nu_1=1$, $\tau=0.0001$, and $T=0.5$, as in Section~\ref{SecNume}. In Figure~\ref{fig:graph}, the color of each node indicates the category to which it belongs. The edges between nodes in the same category are colored the same as the nodes, while edges between nodes in different categories are colored gray.


As shown in Figure~\ref{fig:graph}, there are 42 edges between nodes belonging to the `Fashion' category. This indicates that the orders of products in the `Fashion' category affect each other, which is consistent with the fact that the percentage of customers who bought multiple fashion-related products is the highest among all categories. For example, a customer who bought a suit was more likely to buy a shirt, formal shoes, and a titan watch. On the other hand, a customer who bought jeans was more likely to buy casual shoes, running shoes, and sneakers. Additionally, there are 15 edges from nodes categorized as `Home \& Furniture' to nodes categorized as `Fashion'. This suggests that customers who bought home-related products were also likely to buy fashion-related products. For example, a customer who bought a shoe rack was more likely to buy accessories and shoes, such as a titan watch and formal shoes. All the 57 edges are concentrated in two categories, `Fashion' and `Home \& Furniture', which agrees with the fact that most orders are concentrated in these two categories. 

The directions of edges are associated with the order amounts of products. Precisely, it tends to have directed edges from the less sold products to the more sold products. For example, the best-selling product, a titan watch, has the highest indegree of 11, while the second best-selling product, formal shoes, has an indegree of 8, which is the second largest. The relationship between the directions of edges and the order amounts is intuitive, because a product that is likely to be purchased with other items is likely to have high order amounts. Hence, we can conclude that the estimated graph provides reasonable information regarding the purchasing behaviors of online customers. In other words, it demonstrates that the proposed algorithm can identify legitimate edges in a sparse graph with some hub nodes.

\section{Summary}

\label{SecFuture}

This study proposes the first consistent Bayesian algorithm for learning high-dimensional linear BNs with light and heavy-tailed error distributions. More precisely, this study shows that the proposed algorithm can learn linear Bayesian networks with sub-Gaussian and 4m-th bounded-moment error distributions, with sufficient numbers of samples $n = \Omega( d_M^2 \log p)$ and $n = \Omega(d_M^2 p^{2/m})$, respectively. 
It also shows that the algorithm can outperform frequentist algorithms that rely on different assumptions, but with the same complexity.

However, the proposed algorithm suffers from heavy computational costs and requires careful selection of the hyper-parameters. The hyper-parameters can be chosen by the cross validation, or simply selected by respecting the recommended small spike and tuning parameters as discussed in Sections~\ref{SecAlgo} and~\ref{Secillustration}. The theoretical guarantees of the algorithm are supported through various numerical experiments.  

In future work, it would be interesting to develop faster Bayesian method using the topological layer of a graph as in other DAG model learning algorithms (e.g., \citealp{gao2020polynomial,zhou2021efficient,park2023computationally}). Additionally, it would be important to develop Bayesian approach for other identifiable DAG models, such as nonlinear additive noise models (\citealp{zhang2009causality,hoyer2009nonlinear,mooij2009regression,peters2012identifiability}), non-Gaussian linear BNs (\citealp{shimizu2006linear,zhang2009identifiability}), and count DAG models (\citealp{park2017learning}).

\appendix

\section{Proof of Theorem~\ref{thm001} }

\label{SecProofThm001}

This section proves that Algorithm~\ref{OurAlgorithm} accurately recover the underlying graph with high probability. 
The proof is built upon the prior works of linear BN learning algorithms~\citet{park2020identifiability2,ghoshal2018learning} and theoretical result of BAGUS in \citet{gan2019bayesian}. Here, we restate the proof in our framework. 

Assume that the true ordering satisfying the backward selection condition in Lemma~\ref{lem:identifiability} is $\pi = (\pi_1, \pi_2, ..., \pi_p) = (p,p-1,...,1)$, and hence $\pi_{p+1-r} = r$. For ease of notation, let $\pi_{1:j} = (\pi_1, \pi_2, ..., \pi_j)$. 
For any matrix $A \in \mathbb{R}^{p \times p}$, let $||A ||_{\infty} = \max_{1\le i,j \le p}|[A]_{i,j}|$.
Lastly, $\sigma_{\max}^2 = \max_{1\le j \le p} \sigma_j^2$ is the maximum error variance. 

In this proof, we show that Algorithm~\ref{OurAlgorithm} accurately recover the graph structure if a sample covariance matrix error is sufficiently small using mathematical induction.

\begin{lemma}
	\label{lem001}
	The diagonal entries of the inverse covariance matrix are given as:
	\begin{align*}
		[\Omega]_{k,k} = \frac{1}{\sigma_k^2} + \sum_{l \in \ch(k)}\frac{\beta_{k,l}^2}{\sigma_l^2}.
	\end{align*}
\end{lemma}

\begin{lemma}
	\label{lem002}
	Consider a linear BN~\eqref{eq:LinearSEM} with sub-Gaussian and $4m$-th bounded-moment errors. For any pre-defined constants $C_1>0$ and $C_2 > C_3>0$, assume
	\begin{itemize}
		\item[i)] the prior hyper-parameters $v_0, v_1, \eta$, and $\tau$ satisfy
		\begin{equation*}
			\begin{cases}
				\frac{1}{nv_1} = C_3\frac{1}{p}\frac{1}{1+\epsilon_1}, \frac{1}{nv_0} > C_4\frac{1}{p}\\
				\frac{v_1^2(1-\eta)}{v_0^2\eta} \le \epsilon_1\exp\left[ 2(C_2-C_3)M_{\Gamma}(C_4-C_3)n/p^2\right] , \text{ and } \tau \le C_3\frac{n}{2}\frac{1}{p}
			\end{cases}
		\end{equation*}
		for some constants $\epsilon_1>0, C_4 = C_1+M_{\Sigma}^22(C_1+C_3)M_{\Gamma}+6(C_1+C_3)^2M_{\Gamma}^2M_{\Sigma}^3$,
		\item[ii)] the spectral norm bound $B_0$ satisfies $\frac{1}{k_1} + 2(C_1+C_3)M_{\Gamma} < B_0 < (2nv_0)^\frac{1}{2}$,
		\item[iii)] 
		$\max \left\{2(C_1+C_3)M_{\Gamma} \max \left(3M_{\Sigma}, 3M_{\Gamma}M_{\Sigma}^3, \frac{2}{k_1^2}\right), \frac{2C_3\epsilon_1}{k_1^2p}\right\} \le 1$,
		\item [iv)]
		$\Lambda_{\min}(\Sigma) \ge k_1$.
	\end{itemize}
	If $\|\widehat{\Sigma} - \Sigma \|_{\infty} < C_1\frac{1}{d}$, then
	\begin{align*}
		\|\widehat{\Omega} - \Omega \|_{\infty} < 2(C_1+C_3)M_{\Gamma}\frac{1}{d},
	\end{align*}
	where $d = \max_{r \in \{1,2,\cdots,p\}}\max_{ i \in \{\pi_1,\cdots,\pi_r\} } card\{j : \Omega^{(r)}_{i,j} \neq 0\}$.
\end{lemma}

Assume
\begin{align*}
	\|\widehat{\Sigma} - \Sigma \|_{\infty} < C_1\frac{1}{d}.
\end{align*}	
The last element of the ordering, say $\pi_p$, is achieved by comparing diagonal entries of the MAP estimator $\widehat{\Omega}$ in Equation~\eqref{eq:BAGUS}. More specifically, it is estimated as $\widehat{\pi}_p = \arg \min_{k \in V} [\widehat{\Omega} ]_{k,k}$.
If the following inequality holds, Algorithm~\ref{OurAlgorithm} estimates node 1 as $\widehat{\pi}_p$, so the terminal vertex of the graph is correctly recovered, $\widehat{\pi}_p = \pi_p$.
\begin{align}\label{diag_hat_ineq}
	\min_{ k \in \pi_{1:p-1}} \left([\widehat{\Omega}]_{k,k} - [\widehat{\Omega}]_{1,1}\right) > 0.
\end{align}	
Since
\begin{align*}
	\min_{ k \in \pi_{1:p-1} } \left([\widehat{\Omega}]_{k,k} - [\widehat{\Omega}]_{1,1}\right) &= \min_{ k \in \pi_{1:p-1} } \left\{ \Big([{\Omega}]_{k,k} - [{\Omega}]_{1,1}\Big)-\left([\widehat{\Omega}]_{1,1} - [{\Omega}]_{1,1} \right) + \left([\widehat{\Omega}]_{k,k} - [{\Omega}]_{k,k} \right) \right\} \\
	& \ge \min_{ k \in \pi_{1:p-1} } \left\{ \Big([{\Omega}]_{k,k} - [{\Omega}]_{1,1}\Big)-\left|[\widehat{\Omega}]_{1,1} - [{\Omega}]_{1,1} \right| - \left|[\widehat{\Omega}]_{k,k} - [{\Omega}]_{k,k} \right| \right\}  ,
\end{align*}	
the Inequality \eqref{diag_hat_ineq} holds if
\begin{align}
	\min_{ k \in \pi_{1:p-1} } \left([{\Omega}]_{k,k} - [{\Omega}]_{1,1}\right) &> 4(C_1+C_3)M_{\Gamma}\frac{1}{d} \label{omega_gap1} \\
	\text{ and }\quad \max_{ k \in V } 
	\left|[\widehat{\Omega}]_{k,k} - [{\Omega}]_{k,k}\right| 
	&< 2(C_1+C_3)M_{\Gamma}\frac{1}{d}. \label{omega_gap2}
\end{align}	
Inequality \eqref{omega_gap1} holds due to Assumption~\ref{assu:condvar}, condition in Theorem~\ref{thm001} and Lemma~\ref{lem001}. 
Also, since the constants $C_1,C_2,C_3$ and hyper-parameters $\nu_0,\nu_1,\eta,\tau$ satisfy the conditions i),ii),iii) in Lemma~\ref{lem002} and condition iv) holds by Assumption~\ref{assu:eigen}, we can apply Lemma~\ref{lem002}. Then Inequality \eqref{omega_gap2} holds. 
Hence, Inequality \eqref{diag_hat_ineq} holds, which implies that the last element of the ordering is correctly estimated, i.e., $\widehat{\pi}_p = \pi_p =1$, if $\|\widehat{\Sigma} - \Sigma \|_{\infty} < C_1/ d$. 

\begin{lemma}
	\label{lem003}
	Suppose that Assumption~\ref{assu:eigen}, \ref{assu:condvar}, \ref{assu:min} and conditions i), ii), iii) in Theorem~\ref{thm001} are satisfied. If $\|\widehat{\Sigma} - \Sigma \|_{\infty} < C_1/d$, 
	$$\widehat{S} = supp(\Omega)$$ 
	where $\widehat{S}= \{ (j,k) : p_{jk} \ge T \}$, 
	by choosing appropriate threshold $T$.
\end{lemma}

\begin{lemma}
	\label{lem004}
	If $k$ is a terminal vertex of the graph, then $\pa(k) = supp([\Omega]_{k,*})\setminus \{k\}$, where $[\Omega]_{k,*}$ is the $k$-th row of $\Omega$.
\end{lemma}	
Hence, by Lemmas~\ref{lem003} and \ref{lem004}, if $\|\widehat{\Sigma} - \Sigma \|_{\infty} < C_1/ d$, parents of $\pi_p$ are correctly estimated:
\begin{align*}
	\widehat{\pa}\left(\widehat{\pi}_p\right) = 
	\pa\left(\pi_p\right).
\end{align*}

Now, suppose that we have correctly estimated $\left\{ {\pi}_{p+2-r},  \ldots, {\pi}_p \right\}$ for some $r= 2,3 ,\ldots,p-1 $, i.e.,  $\left\{   \widehat{\pi}_{p+2-r}  , \ldots, \widehat{\pi}_p \right\} = \left\{ {\pi}_{p+2-r}  , \ldots,   {\pi}_p \right\}$. 
Then if $\|\widehat{\Sigma} - \Sigma \|_{\infty} < C_1/ d$, the following inequality holds:
\begin{align*}
	\|\widehat{\Sigma}^{(r)} - {\Sigma}^{(r)}\|_{\infty} \le \|\widehat{\Sigma} - {\Sigma}\|_{\infty} \le C_1\frac{1}{d},
\end{align*}
where $\widehat{\Sigma}^{(r)}$ and ${\Sigma}^{(r)}$ are the sample and true covariance matrix of $( X_{\pi_1}, \ldots,  X_{\pi_{p+1-r}} )$.	
Also, we can derive the minimum eigenvalue of ${\Sigma}^{(r)}$ by the following inequality:
\begin{align}\label{SubCovEig}
	\Lambda_{\min}({\Sigma}^{(r)}) = \min_{||x||_2=1, x \in \mathbb{R}^{p+1-r}} x^\top \Sigma^{(r)} x
	\ge \min_{||x||_2=1, x \in \mathbb{R}^{p}} x^\top \Sigma x
	=\Lambda_{\min}(\Sigma) \ge k_1
\end{align}
The $(p+1-r)$-th element of the ordering, say $\pi_{p+1-r}$, is achieved by comparing diagonal entries of the estimated inverse covariance matrix of $( X_{\pi_1}, \ldots,  X_{\pi_{p+1-r}} )$,
denoted as $\widehat{\Omega}^{(r)}$. It is correctly estimated if the following inequality holds:
\begin{align}\label{diag_hat_ineq_r}
	\min_{ k \in \pi_{1:p-r} } \left([\widehat{\Omega}^{(r)}]_{k,k} - [\widehat{\Omega}^{(r)}]_{r,r}\right) > 0.
\end{align}	
By similar arguments used to show Inequality~\eqref{diag_hat_ineq}, the above inequality holds if 
\begin{align*}
	\min_{ k \in \pi_{1:p-r} } \left([{\Omega}^{(r)}]_{k,k} - [{\Omega}^{(r)}]_{r,r}\right)  &> 4(C_1+C_3)M_{\Gamma^{(r)}}\frac{1}{d}  \\
	\text{ and }\quad  \max_{ k \in \pi_{1:p+1-r} } \left([\widehat{\Omega}^{(r)}]_{k,k} - [{\Omega}^{(r)}]_{k,k}\right) &< 2(C_1+C_3)M_{\Gamma^{(r)}}\frac{1}{d}.
\end{align*}	
The first inequality holds due to Assumption~\ref{assu:condvar}, condition in Theorem~\ref{thm001} and applying Lemma~\ref{lem001} to ${\Omega}^{(r)}$. 	
Also, since the constants $C_1,C_2,C_3$ and hyper-parameters $\nu_0,\nu_1,\eta,\tau$ satisfy the conditions i),ii),iii) in Lemma~\ref{lem002}, condition iv) holds by \eqref{SubCovEig} and $\|\widehat{\Sigma}^{(r)} - {\Sigma}^{(r)}\|_{\infty} \le C_1\frac{1}{d}$, we can apply Lemma~\ref{lem002} to $\widehat{\Sigma}^{(r)}$ and $\widehat{\Omega}^{(r)}$. Then the second inequality holds. 	
Thus, if $\|\widehat{\Sigma} - \Sigma \|_{\infty} < C_1/ d$, Inequality \eqref{diag_hat_ineq_r} holds, which implies  $\widehat{\pi}_{p+1-r} = {\pi}_{p+1-r}= r$. 
Again by Lemmas \ref{lem003} and \ref{lem004}, if $\|\widehat{\Sigma} - \Sigma \|_{\infty} < C_1/ d$, parents of $\pi_{p+1-r}$ are correctly estimated:
\begin{align*}
	\widehat{\pa}\left(\widehat{\pi}_{p+1-r}\right) = \widehat{\pa}\left(\pi_{p+1-r}\right) = \pa\left(\pi_{p+1-r}\right).
\end{align*}

Therefore, by the mathematical induction, $\widehat{G}=G$ holds 
if $\|\widehat{\Sigma} - \Sigma \|_{\infty} < C_1/ d$. 	
Using this result, the following inequality holds:
\begin{align*}
	& \Pr\left(\widehat{G} = G\right) \\
	\ge & \Pr\left( \|\widehat{\Sigma} - \Sigma \|_{\infty} < C_1\frac{1}{d}\right) \\
	= & 1 - \Pr\left( \|\widehat{\Sigma} - \Sigma \|_{\infty} \ge C_1\frac{1}{d}\right) \\
	= & 1 - \Pr\left(\cup_{i,j \in \{1,2, ... ,p\} } \left[| [\widehat{\Sigma}]_{i,j} -  [\Sigma]_{i,j}   | \ge C_1\frac{1}{d}\right]\right) \\
	\ge & 1 - \sum_{i,j \in \{1,2, ... ,p\}} \Pr\left(|   [\widehat{\Sigma}]_{i,j} -  [\Sigma]_{i,j}   | \ge C_1\frac{1}{d}\right)
\end{align*}

\begin{lemma}[Error Bound for the Sample Covariance Matrix]
	\label{lem005} Consider a random vector $(X_j)_{j = 1}^{p}$ whose covariance matrix is $\Sigma \in \mathbb{R}^{p\times p}$. 
	\begin{itemize}
		\item Lemma 1 of \citet{ravikumar2011high}: 
		When $(X_j)_{j = 1}^{p}$ follows a sub-Gaussian distribution,
		\begin{align*}
			\Pr \bigg( \big|  [\widehat{\Sigma}]_{j,k} -  [\Sigma]_{j,k}  \big| \geq \zeta \bigg) \leq 4 \cdot \exp\left( \frac{-n \zeta^2 }{ 128( 1+ 4 s_{\max}^2 )  \max_{j} ([\Sigma]_{j,j})^2 } \right), 
		\end{align*}
		for all $\zeta \in (0, \max_{j} [\Sigma]_{j,j} 8( 1 + 4 s_{\max}^2 ) )$.
		\item Lemma 2 of \citet{ravikumar2011high}:
		When $(X_j)_{j = 1}^{p}$ follows a 4m-th bounded-moment distribution,
		\begin{align*}
			\Pr \bigg(   \big|  [\widehat{\Sigma}]_{j,k} -  [\Sigma]_{j,k}  \big| \geq \zeta \bigg) \leq 4 \cdot \frac{ 2^{2m} \max_{j} ( [\Sigma]_{j,j})^{2m} C_m(K_{\max} + 1) }{ n^m \zeta^{2m} }, 
		\end{align*}
		where $C_m$ is a constant depending only on $m$. 
	\end{itemize}
\end{lemma}

\begin{lemma}
	\label{lem006} For a linear BN, the following inequality holds between column sparsity $d$ and maximum degree of the moralized graph $d_M$:
	\begin{align*}
		d \le d_M+1.
	\end{align*}
\end{lemma}

Applying Lemmas \ref{lem005} and \ref{lem006}, for a sub-Gaussian linear BN \eqref{subGaussian}, 
\begin{align*}
	& \Pr\left(\widehat{G} = G\right) \\
	\ge & 1 - \sum_{i,j \in \{1,2, ... ,p\}} \Pr\left(| [\widehat{\Sigma}]_{i,j} -  [\Sigma]_{i,j} | \ge C_1\frac{1}{d}\right) \\
	\ge & 1-4p^2\exp\left(  - \frac{ C_1^2 }{128(1+4 s_{\max}^2 )\max_{j}\left(  [\Sigma]_{j,j}\right)^2}\frac{n}{d^2}\right) \\
	\ge &
	1-4p^2\exp\left( - \frac{ C_1^2  }{128(1+4 s_{\max}^2)\max_{j}\left( [\Sigma]_{j,j}\right)^2}\frac{n}{(d_M+1)^2}\right).
\end{align*}

Applying Lemma~\ref{lem005} and \ref{lem006}, for a $4m$-th bounded-moment linear BN \eqref{4mbdd}, 
\begin{align*}
	& \Pr\left(\widehat{G} = G\right) \\
	\ge & 1 - \sum_{i,j \in \{1,2, ... ,p\}} \Pr\left(| [\widehat{\Sigma}]_{i,j} -  [\Sigma]_{i,j} | \ge C_1\frac{1}{d}\right) \\
	\ge & 1-4p^2\frac{2^{2m}\max_{j}\left( [\Sigma]_{j,j}\right)^{2m}C_m(K_{\max}+1)d^{2m}}{n^mC_1^{2m}} \\
	\ge &
	1-4p^2\frac{2^{2m}\max_{j}\left( [\Sigma]_{j,j}\right)^{2m}C_m(K_{\max}+1)(d_M+1)^{2m}}{n^mC_1^{2m}}.
\end{align*}	
\begin{flushright}
	$\blacksquare$
\end{flushright}

\section{ Proofs of lemmas }

\subsection{Proof of Lemma~\ref{lem001}}
\textbf{Lemma~\ref{lem001} }
The diagonal entries of the inverse covariance matrix are given as:
\begin{align*}
	[\Omega]_{k,k} = \frac{1}{\sigma_k^2} + \sum_{l \in \ch(k)}\frac{ \beta_{k,l}^2 }{\sigma_l^2}.
\end{align*}

\begin{proof}
	By \eqref{eq:MatrixSEMSigma}, the inverse covariance matrix can be expressed as
	\begin{align*}
		\Omega = (I_p-B)^\top(\Sigma^\epsilon)^{-1}(I_p-B).
	\end{align*}
	So the diagonal entries $[\Omega]_{k,k}$ are calculated as following:
	\begin{align*}
		[\Omega]_{k,k} = \frac{1}{\sigma_k^2} + \sum_{l \in \ch(k)}\frac{ \beta_{k,l}^2 }{\sigma_l^2}.
	\end{align*} 
\end{proof}

\subsection{Proof of Lemma~\ref{lem002}}
\textbf{Lemma~\ref{lem002} }
Consider a linear BN~\eqref{eq:LinearSEM} with sub-Gaussian and $4m$-th bounded-moment errors. For any pre-defined constants $C_1>0$ and $C_2 > C_3>0$, assume
\begin{itemize}
	\item[i)] the prior hyper-parameters $v_0, v_1, \eta$, and $\tau$ satisfy
	\begin{equation*}
		\begin{cases}
			\frac{1}{nv_1} = C_3\frac{1}{p}\frac{1}{\epsilon_1}, \frac{1}{nv_0} > C_4\frac{1}{p}\\
			\frac{v_1^2(1-\eta)}{v_0^2\eta} \le \epsilon_1\exp\left[ 2(C_2-C_3)M_{\Gamma}(C_4-C_3)n/p^2\right] , \text{ and } \tau \le C_3\frac{n}{2}\frac{1}{p}
		\end{cases}
	\end{equation*}
	for some constants $\epsilon_1>0, C_4 = C_1+M_{\Sigma}^22(C_1+C_3)M_{\Gamma}+6(C_1+C_3)^2M_{\Gamma}^2M_{\Sigma}^3$,
	\item[ii)] the spectral norm bound $B_0$ satisfies $\frac{1}{k_1} + 2(C_1+C_3)M_{\Gamma} < B_0 < (2nv_0)^\frac{1}{2}$,
	\item[iii)] 
	$\max \left\{2(C_1+C_3)M_{\Gamma} \max \left(3M_{\Sigma}, 3M_{\Gamma}M_{\Sigma}^3, \frac{2}{k_1^2}\right), \frac{2C_3\epsilon_1}{k_1^2p}\right\} \le 1$,
	\item [iv)]
	$\Lambda_{\min}(\Sigma) \ge k_1$.
\end{itemize}
If $\|\widehat{\Sigma} - \Sigma \|_{\infty} < C_1\frac{1}{d}$, then
\begin{align*}
	\|\widehat{\Omega} - \Omega \|_{\infty} < 2(C_1+C_3)M_{\Gamma}\frac{1}{d}.
\end{align*}

\begin{proof}
	Before presenting the proof, we will define two penalize functions $pen_{SS}(\theta), pen_1(\theta)$ as following:
	\begin{align*}
		&pen_{SS}(\theta) = -\log\left[\left(\frac{\eta}{2\nu_1}\right)e^{-\frac{|\theta|}{\nu_1}} + \left(\frac{1-\eta}{2\nu_0}\right)e^{-\frac{|\theta|}{\nu_0}}\right],\\
		&pen_1(\theta) = \tau|\theta|.
	\end{align*}
	By the definition, we can find bounds of the first and second derivatives of $pen_{SS}(\delta)$, respectively.
	\begin{itemize}
		\item Bound of the first derivative of $pen_{SS}(\delta)$:
		
		Using equation (5) in the appendix of \citet{gan2019bayesian},
		the following inequality holds:
		\begin{align*}
			\frac{1}{n}|pen'_{SS}(\delta)| < \frac{1}{n\nu_1}\left(1+\frac{\frac{\nu_1^2(1-\eta)}{\nu_0^2\eta}}{e^{\frac{|\delta|}{\nu_0}-\frac{|\delta|}{\nu_1}}}\right).
		\end{align*}
		With the conditions in this lemma and letting $\nu_1^2(1-\eta)/(\nu_0^2\eta) = \xi \exp\left[ \psi(C_4-C_3)n/p^2\right]$ where $\xi < \epsilon_1$, 
		\begin{align*}
			\frac{\frac{\nu_1^2(1-\eta)}{\nu_0^2\eta}}{e^{\frac{|\delta|}{\nu_0}-\frac{|\delta|}{\nu_1}}} \le \frac{\xi \exp\left[ \psi(C_4-C_3)n/p^2\right]}{\exp\left[\psi(C_4-C_3)n/p^2\right]} \le \xi,
		\end{align*} 
		when $|\delta| \ge \psi/p$. So we can get the bound of $\frac{1}{n}|pen'_{SS}(\delta)|$ as
		\begin{align*}
			\frac{1}{n}|pen'_{SS}(\delta)| < \frac{1}{n\nu_1}\left(1+\frac{\frac{\nu_1^2(1-\eta)}{\nu_0^2\eta}}{e^{\frac{|\delta|}{\nu_0}-\frac{|\delta|}{\nu_1}}}\right) \le
			C_3\frac{1}{p}\frac{1}{1+\epsilon_1}(1+\xi) < C_3\frac{1}{p} \le C_3\frac{1}{d}.
		\end{align*}
		\item Bound of the second derivative of $pen_{SS}(\delta)$:
		
		Using equation (7) in the appendix of \citet{gan2019bayesian}
		and since the condition iii) implies $\frac{1}{p} \le \frac{k_1^2}{2C_3\epsilon_1}$, the following inequality holds:
		\begin{align*}
			\frac{1}{2n}|pen''_{SS}(\delta)| < \frac{\xi}{2n\nu_1} < \frac{C_3}{2}\xi\frac{1}{p} < \frac{C_3}{2}\epsilon_1\frac{1}{p} \le \frac{1}{4}k_1^2.
		\end{align*}
	\end{itemize}
	Using the bounds of the first and second derivatives of $pen_{SS}(\delta)$ derived above, this lemma can be proved as same as Theorem A in the appendix of \citet{gan2019bayesian}
	changing $\sqrt{\frac{\log p}{n}}$ to $\frac{1}{d}$. 
	Also, we can omit the sample size bound $n \ge M^2 \log p$ in 
	Theorem A in the appendix of \citet{gan2019bayesian}
	so this lemma can be used more freely without considering the condition of $n$ and $p$.
\end{proof}

\subsection{Proof of Lemma~\ref{lem003}}
\textbf{Lemma~\ref{lem003} }
Suppose that Assumption~\ref{assu:eigen}, \ref{assu:condvar}, \ref{assu:min} and conditions i), ii), iii) in Theorem~\ref{thm001} are satisfied. If $\|\widehat{\Sigma} - \Sigma \|_{\infty} < C_1/d$, 
$$\widehat{S} = supp(\Omega)$$ 
where $\widehat{S}= \{ (j,k) : p_{jk} \ge T \}$, 
by choosing appropriate threshold $T$.

\begin{proof}
	Set the threshold $T$ satisfies the following range:
	\begin{align}\label{ThresRange}
		\log\left(\frac{T}{1-T}\right) \in \log\left(\frac{\nu_0\eta}{\nu_1(1-\eta)}\right)+ \left(0, \left(\theta_{min}-2(C_1+C_3)M_{\Gamma}\frac{1}{d}\right)\left(\frac{1}{\nu_0}-\frac{1}{\nu_1}\right)\right).
	\end{align}
	
	Equation (9) of \citet{gan2019bayesian} is equivalent to
	\begin{align*}
		&\log\left(\frac{p_{ij}}{1-p_{ij}}\right) \\ 
		= & -\log\left(\frac{\nu_1(1-\eta)}{\nu_0\eta}\right) + |\widehat{\theta}_{i,j}|\left(\frac{1}{\nu_0}-\frac{1}{\nu_1}\right) \\
		= & \log\left(\frac{\nu_0\eta}{\nu_1(1-\eta)}\right)+ |\widehat{\theta}_{i,j}|\left(\frac{1}{\nu_0}-\frac{1}{\nu_1}\right).
	\end{align*}
	
	To check a support of the estimated inverse covariance matrix is correctly estimated, we consider two cases which an element of the true inverse covariance matrix is zero or not, $\theta_{i,j} = 0$ and $\theta_{i,j} \neq 0$.
	
	\begin{itemize}
		\item When $\theta_{i,j}=0$, by constructor in the proof of Lemma~\ref{lem002}, $\widehat{\theta}_{i,j}=0$.
		Hence, the following inequality holds using the lower bound of Range \eqref{ThresRange}:
		\begin{align*}
			\log\left(\frac{p_{ij}}{1-p_{ij}}\right)
			= & \log\left(\frac{\nu_0\eta}{\nu_1(1-\eta)}\right) \\
			< & \log\left(\frac{T}{1-T}\right).
		\end{align*}
		
		This inequality implies $p_{i,j} < T$, so the zero entries are recovered correctly.
		
		\item When $\theta_{i,j} \neq 0$, applying Lemma~\ref{lem002}, the following inequality holds:
		\begin{align*}
			|\widehat{\theta}_{i,j}| > \theta_{min} - 2(C_1+C_3)M_{\Gamma}\frac{1}{d} > 0.
		\end{align*}
		
		Then the following inequality holds using the upper bound of Range \eqref{ThresRange}:
		\begin{align*}
			\log\left(\frac{p_{ij}}{1-p_{ij}}\right)
			= & \log\left(\frac{\nu_0\eta}{\nu_1(1-\eta)}\right) + |\widehat{\theta}_{i,j}|\left(\frac{1}{\nu_0}-\frac{1}{\nu_1}\right) \\
			> & \log\left(\frac{\nu_0\eta}{\nu_1(1-\eta)}\right) + \left(\theta_{min}-2(C_1+C_3)M_{\Gamma}\frac{1}{d}\right)\left(\frac{1}{\nu_0}-\frac{1}{\nu_1}\right) \\
			> & \log\left(\frac{T}{1-T}\right).
		\end{align*}
		
		This inequality implies $p_{i,j} > T$, so the nonzero entries are recovered correctly.	
	\end{itemize}
	
	Therefore, a support of the true inverse covariance matrix is correctly recovered, $supp(\widehat{\Omega}) = supp(\Omega)$.
\end{proof}

\subsection{Proof of Lemma~\ref{lem004}}
\textbf{Lemma~\ref{lem004} }
If $k$ is a terminal vertex of the graph, then $\pa(k) = supp(\Omega_{k,*})\setminus \{k\}$.

\begin{proof}
	By \eqref{eq:MatrixSEMSigma}, the inverse covariance matrix can be expressed as
	\begin{align*}
		\Omega = (I_p-B)^\top(\Sigma^\epsilon)^{-1}(I_p-B).
	\end{align*}
	So the entries $[\Omega]_{k,j}$ are calculated as following:
	\begin{equation}
		\label{eq:invcovent}
		[\Omega]_{k,j} = -\frac{1}{\sigma_j^2}\beta_{k,j} -\frac{1}{\sigma_k^2}\beta_{j,k} + \sum_{l \in \ch(k) \cap \ch(j)}\frac{ \beta_{k,l}\beta_{j,l} }{\sigma_l^2}.
	\end{equation}
	If $k$ is a terminal vertex of the graph, $\beta_{k,j}=0$ for all $j \in V\setminus\{k\}$. Then the following equation holds:
	\begin{align*}
		[\Omega]_{k,j} = -\frac{1}{\sigma_k^2}\beta_{j,k}.
	\end{align*}
	Since $\pa(k) = \left\{j \in V | \beta_{j,k} \neq 0\right\} $, the parent set of $k$ can be found using the support set of the $k$-th row of the inverse covariance matrix as following:
	\begin{align*}
		\pa(k) = supp(\Omega_{k,*})\setminus \{k\}
	\end{align*}
\end{proof}

\subsection{Proof of Lemma~\ref{lem005}}
\textbf{Lemma~\ref{lem005} }[Error Bound for the Sample Covariance Matrix]
Consider a random vector $(X_j)_{j = 1}^{p}$ and suppose that its covariance matrix is $\Sigma$. 
\begin{itemize}
	\item Lemma 1 of \citet{ravikumar2011high}: 
	When $(X_j)_{j = 1}^{p}$ follows a sub-Gaussian distribution,
	\begin{align*}
		\Pr \bigg( \big|  [\widehat{\Sigma}]_{j,k} -  [\Sigma]_{j,k}  \big| \geq \zeta \bigg) \leq 4 \cdot \exp\left( \frac{-n \zeta^2 }{ 128( 1+ 4 s_{\max}^2 )  \max_{j} ([\Sigma]_{j,j})^2 } \right), 
	\end{align*}
	for all $\zeta \in (0, \max_{j} [\Sigma]_{j,j} 8( 1 + 4 s_{\max}^2 ) )$.
	\item Lemma 2 of \citet{ravikumar2011high}:
	When $(X_j)_{j = 1}^{p}$ follows a 4m-th bounded-moment distribution,
	\begin{align*}
		\Pr \bigg(   \big|  [\widehat{\Sigma}]_{j,k} -  [\Sigma]_{j,k}  \big| \geq \zeta \bigg) \leq 4 \cdot \frac{ 2^{2m} \max_{j} ( [\Sigma]_{j,j})^{2m} C_m(K_{\max} + 1) }{ n^m \zeta^{2m} }, 
	\end{align*}
	where $C_m$ is a constant depending only on $m$. 
\end{itemize}

Since it is the same as Lemmas 1 and 2 from \citet{ravikumar2011high}, we omit the proof. 

\subsection{Proof of Lemma~\ref{lem006}}
\textbf{Lemma~\ref{lem006} }
For a linear BN, the following inequality holds between column sparsity $d$ and maximum degree of the moralized graph $d_M$:
\begin{align*}
	d \le d_M+1.
\end{align*}

\begin{proof}
	If two nodes $j,k$ is not connected in the moralized graph, it means that $[B]_{j,k}=[B]_{k,j}=0$ and there are no common child between $j$ and $k$. Then by \eqref{eq:invcovent}, it can be shown that $[\Omega]_{k,j}=0$. So if $[\Omega]_{j,k} \neq 0$, nodes $j$ and $k$ are connected in the moralized graph.
	Hence, the following inequality holds:
	\begin{align*}
		d &= \max_{i =1,2,\cdots p} card\{j: [\Omega]_{i,j} \neq 0\}\\ 
		&=1+\max_{i =1,2,\cdots p} card\{j \neq i: [\Omega]_{i,j} \neq 0\}\\
		&\le 1+\max_{i =1,2,\cdots p} card\{j: (i,j) \in M(G)\}\\
		&= d_M+1,
	\end{align*}
	where $M(G)$ is the moralized graph of $G$.
\end{proof}
	
	\vskip 0.2in
	
	\bibliography{BayBayNet_reference}
	
\end{document}